\documentclass[final,3p,authoryear]{elsarticle}

\usepackage{algorithm}
\usepackage{algpseudocode}
\usepackage{mathtools}

\usepackage{subcaption}
\usepackage{verbatim} %multiline comment
\usepackage{dcolumn}
\usepackage{multirow}
\usepackage[table]{xcolor}
\usepackage{color,soul}
\usepackage[normalem]{ulem} % creating strickthrough lines
\usepackage{soul}

\emergencystretch=\maxdimen
\DeclareCaptionLabelFormat{bf-shortfig}{\textbf{#2.}}
\DeclareCaptionLabelFormat{bf-fulltable}{\text{#2}}
\usepackage[colorlinks=true,urlcolor=blue,citecolor=blue,linkcolor=blue,bookmarks=true]{hyperref}
\usepackage{graphicx}
\usepackage{float}

\begin{document}
\begin{frontmatter}
\title{Genetic Neural Architecture Search for automatic assessment of human sperm images}
\author[label1]{Erfan Miahi}
\ead{mhi.erfan1@gmail.com}

\author[label1]{SeyedAbolghasem Mirroshandel\corref{cor1}}
\address[label1]{Department of Computer Engineering, University of Guilan, Rasht, Iran}
\ead{mirroshandel@guilan.ac.ir}
\cortext[cor1]{Corresponding author. Tel.: +98 13 33690274 (Ext. 3193); fax: +98 1333690271}

\author[label2]{Alexis Nasr}
\address[label2]{Laboratoire d\' Informatique et Systemes, Aix Marseille Universit\'e, Marseille, France}
\ead{alexis.nasr@univ-amu.fr}

\begin{abstract}
Male infertility is a disease that affects approximately 7\% of men. Sperm morphology analysis (SMA) is one of the main diagnosis methods for this problem. However, manual SMA is an inexact, subjective, non-reproducible, and hard to teach process. Therefore, in this paper, we introduce a novel automatic SMA technique that is based on the neural architecture search algorithm, named Genetic Neural Architecture Search (GeNAS). For this purpose, we used a collection of images termed MHSMA dataset, which contains $1,540$ sperm images that have been collected from $235$ patients with infertility problems. In detail, GeNAS consists of a special genetic algorithm that acts as a meta-controller which explores the constrained search space of plain convolutional neural network architectures.
Every individual of this genetic algorithm is a convolutional neural network trained to predict morphological deformities in different segments of human sperm (head, vacuole, and acrosome). The fitness of each individual is calculated by a novel proposed method, {named GeNAS Weighting Factor (GeNAS-WF)}. This technique is specially designed to evaluate the fitness of neural networks which, during their learning process, validation accuracy highly fluctuates. To speed up the algorithm, a hashing method is practiced to save each trained neural architecture fitness, so we could reuse them during fitness evaluation. In terms of running time and computational power, our proposed architecture search method is far more efficient than most of the other existing neural architecture search algorithms. Moreover, whereas most of the existing neural architecture search algorithms are designed to work well with well-prepared benchmark datasets, the overall paradigm of GeNAS is specially designed to address the challenges of real-world datasets, particularly shortage of data and class imbalance. In our experiments, the best neural architecture found by GeNAS has reached an accuracy of 91.66\%, 77.33\%, and 77.66\% in the vacuole, head, and acrosome abnormality detection, respectively. 
In comparison to other proposed algorithms for MHSMA dataset, GeNAS achieved state-of-the-art results.

\end{abstract}

\end{frontmatter}

\section{Introduction}

Approximately 15\% of couples suffer from infertility, which is in 30 to 40\% of the cases due to the male sperm abnormalities~\citep{isidori2005treatment,stouffs2008there}. One of the key methods for male infertility diagnosis is sperm morphology analysis (SMA), which consists of classifying sperm head, vacuole, acrosome, and tail as normal or abnormal. However, manual SMA suffers from several flaws. {It is inexact, subjective, non-reproducible, and hard to teach process.}  Therefore, in the past couple of years, several computer-based algorithms have been proposed to address these limitations and automate this process ~\citep{javadi2019novel, riordon2019deep, ghasemian2015efficient}.

Automating the SMA process can be seen as a computer vision classification problem, which aims to classify sperm's head, vacuole, acrosome, and tail. Moreover, in recent years, Convolutional Neural Network (CNN) algorithms became the state-of-the-art technique in solving a lot of computer vision tasks
\citep{voulodimos2018deep}, especially in the area of image classification \citep{krizhevsky2012imagenet, SzegedyIV16, SuganumaSN17}. As a result, these accomplishments make them a natural choice for addressing the SMA. Nevertheless,
usually, a high price should be paid for using CNNs since they exhibit complex architectures and need a large amount of expert knowledge and computer resources to be tuned. This is where Neural Architecture Search (NAS) algorithms come forward and attempt to overcome these challenges \citep{he2019automl}. 
These algorithms can be described as techniques that strive to automate the process of designing neural network architectures. They usually try to achieve this purpose by adopting an efficient search algorithm, such as Evolutionary Algorithms~\citep{real2017large} or Reinforcement Learning~\citep{liu2017hierarchical}, to search for optimal neural network architecture. However, in the past few years, the studies on designing NAS algorithms have confronted with several challenges.

The early proposed methods in the NAS domain were highly computationally expensive. In other words, they employed hundreds of GPU-hours to discover a near-optimal neural architecture. Nonetheless, more recent algorithms attempted to tackle this challenge by employing different techniques, namely network morphism \citep{jin2018efficient}, and predicting neural architecture accuracy \citep{baker2016designing}. However, just a small number of these techniques have succeeded to effectively reduce computational cost.

Furthermore, many of the proposed NAS techniques have been evaluated on benchmark datasets, such as ImageNet~\citep{5206848} and CIFAR-10~\citep{krizhevsky2009learning}. The main advantage of this is that the Machine Learning (ML) researchers can have an integrated testbed to compare the performance of their algorithms. However, the central disadvantages of these could be that proposed algorithms could overfit to predefined standard datasets. To clarify, the problem with these datasets is that they do not represent real-world datasets in industry, medicine, and economy. Such datasets frequently are imbalanced and contain low-quality images with noisy labels. They also suffer from the shortage of data, whereas standard datasets are well-prepared and do not experience these limitations.

In this paper, we adopted a dataset named Modified Human Sperm Morphology Analysis dataset (MSHMA), which contains the features often seen in practical datasets, especially the ones in the field of medicine \citep{javadi2019novel}. In detail, the classes are highly imbalanced, particularly in the vacuole label; the number of images is not adequate; images are non-stained and noisy.

The algorithm proposed in this paper, termed Genetic Neural Architecture Search (GeNAS), can be seen as a framework that employs a customized genetic algorithm to search for an optimal neural architecture. In detail, the population of this genetic algorithm comprises chromosomes, which each one represents a CNN architecture. At first, the population will be initialized with random values and lengths. After fitness calculation, special crossover, mutation, and tournament selection operators will be applied to the population, and this step will be repeated for a specific number of iterations. In this genetic algorithm, while crossover operation explores the depth of neural architectures by combining the parents' genomes, mutation explores the search space of genes' values (e.g., filter-size and stride-size of CNN's architectures). Moreover, the fitness of each individual will be computed using a special technique named GeNAS Weighting Factor (GeNAS-WF), which mainly designed to address the difficulties caused by sampling and data augmentation methods during training. After the genetic algorithm discovers the optimal neural architecture, we train the selected architecture from scratch with different settings. For instance, we increase the number of iterations and change our early stopping technique.

% \hl{The algorithm proposed in this paper, that we call Genetic Neural Architecture Search (GeNAS), consists of a genetic algorithm for searching through architectures of CNNs.} In this framework, an individual is a specific 
% CNN architecture (its hyper-parameters and layers' structure) that is described by a 
% string{,} which constitutes the genome of the individual. The initial population is made of randomly 
% generated genomes of different lengths{, which will be combined} via three genetic operations: tournament
% selection, crossover, and mutation. The crossover operation explores the depth of neural 
% architectures by combining the parents genomes{,} while mutation explores the search space of filter-size
% and stride-size of layers of each neural network by randomly selecting new genome values for 
% each individual. {The fitness of an individual is computed by training it on the training set 
% and computing its accuracy on the validation set. After each iteration, utilizing the obtained 
% validation accuracies by our proposed weighting technique, the final fitness value
% is computed.}

Our experiments show that GeNAS can find CNN architectures that are capable of reaching state-of-the-art accuracy, precision, and $f_{0.5}$ score on the head, acrosome, and vacuole classification for MHSMA dataset~\citep{javadi2019novel}, with fewer parameters and layers, compared to hand-designed models. Moreover, our proposed method works automatically without any human intervention.

The salient features of GeNAS are:

\begin{enumerate}

\item A neural architecture encoding which can explore an optimal constrained search 
space of CNN architectures. 

\item The first and only NAS algorithm for addressing the sperm abnormality detection problem

\item A crossover operation for exploring the depth of neural architecture

\item {A hashing method, for saving pairs of architecture and fitness of each chromosome; and then, reusing them in the fitness evaluation stage to speed up the algorithm}

\item The ability to find the optimal architecture with just 1 Nvidia GPU in less 
than 10 days
    
\item A pruning algorithm during genotype to phenotype conversion to prevent phenotype
(neural architecture) from having negative output height and weight values.

\item A new fitness computation method called GeNAS-WS which is specially designed to 
work with noisy, low quality, and imbalanced datasets
   
\item A neural architecture search algorithm specially designed to work with a challenging dataset which is highly imbalanced; that does not have enough training examples; which images are non-stained and  noisy, which details are not clear and recorded with a low-magnification microscope
\end{enumerate}

The paper is structured as follows: in section~\ref{sec:RelatedWork}, {previous} neural architecture search and sperm assessment methods are introduced. Our proposed algorithm is presented in details in section~\ref{GeNAS}. In the fourth section, MHSMA dataset, augmentation technique, our sampling method, and the results of our experiments both on Random Search and GeNAS are reported and compared to other results on this dataset. In the last section, the conclusion of our proposed method and its results on MHSMA dataset are summarized.

\section{Related Work}
\label{sec:RelatedWork}

Existing studies in the NAS domain and automatic sperm processing are reviewed in this section. All the proposed NAS algorithms can be broken down into three main elements: search space, performance estimation strategy, search strategy \citep{elsken2018neural}. We will elaborate on these components in the following subsection and conclude with a discussion on existing automated sperm processing techniques.

\subsection{Neural Architecture Search Algorithms}
\label{neural_architecture_search}

It is more than three decades that algorithms designed to discover an optimal neural architecture exist. The first of such algorithms employed evolutionary techniques, more precisely genetic algorithm, to search for the best neural architecture and its weights~\citep{miller1989designing, kitano1990designing, schaffer1992combinations, stanley2002evolving}. However, the term NAS just began to show up in recent years. After the publication of \cite{zoph2016neural}, this term became popular, and, since then, a lot of machine learning researchers started to doing research in this field. The main motivation of these researchers is to overcome the limitations which come with hand-designing neural network architectures. In detail, such architectures need a lot of time and expert knowledge to be designed. Moreover, since the number of hyperparameters representing new architectures has grown exponentially by the advancement in the field of deep learning, designing a good architecture became a hard task even for experts. All these proposed algorithms can be represented by three elements, which are discussed in the subsequent paragraphs.

First, the search space can differ depending on the algorithms. By search space, we mean the structure, number, and properties of neural network layers and their connections that the algorithms plan to search through. For instance, for computer vision tasks, while several algorithms attempted to search through plain convolutional neural networks \citep{cai2018efficient}, others tried to explore an expanded version of this search space, which can incorporate skip-connections \citep{real2018regularized}. Furthermore, whereas some of these algorithms aimed to search for the whole architecture \citep{baker2016designing}, others strived to search for a convolutional block (i.e., motifs) \citep{zhong2018practical, zoph2017learning}. In more detail, in the latter version, convolutional blocks are smaller components of the whole CNN, and they will be repeated in a special way to construct the whole architecture.

Admitting all this, different sizes of the search space have their own pros and cons. To clarify, while choosing a large and complex search space can increase the novelty of the discovered architecture, it demands high computational power and time. By considering this, to reduce the computational cost, the search space of GeNAS can only consist of components of plain convolutional neural networks.

The second element is the search strategy. The search strategy can be described as the technique employed to explore the search space in a way that can discover the optimal or near-optimal neural architecture in terms of performance. In the past few years, a lot of techniques, such as Reinforcement Learning~\citep{liu2017hierarchical, real2018regularized, pham2018efficient, cai2018efficient}, Bayesian Optimization~\citep{snoek2012practical, shahriari2016taking, snoek2015scalable, kandasamy2018neural}, Evolutionary Algorithm~\citep{real2017large, xie2017genetic, stanley2002evolving, meyerevolving, miller1989designing}, and gradient-based methods \citep{liu2018darts}, have been used as the search strategy. Nonetheless, among all these approaches, evolutionary algorithms and reinforcement learning are the popular ones. 

The first approach which popularized the NAS domain employed a reinforcement learning algorithm \citep{zoph2016neural}. Accuracy on the validation set was considered as the reward value of this technique. By using this reward and policy iteration algorithm as the main policy, a Recurrent Neural Network (RNN) was trained to generate a variable-length string. This variable-length string describes the structure and connectivity of neural architecture. As a result, while this method succeeded in improving the accuracy on datasets, such as CIFAR10, it was highly computationally expensive (i.e., used 800 GPUs). Hence, it is not possible to use this technique in practice, especially for small companies and individual researchers. Another line of research applied an evolutionary algorithm~\citep{real2017large} for a classification task on CIFAR10 and CIFAR100. This algorithm follows a simple evolutionary algorithm paradigm that first initializes the population and then attempts to improve this population through crossover and mutation. At last, whereas this algorithm achieved a competitive result compared to reinforcement learning and random search algorithms on CIFAR10, it employed a vast amount of computational power: 250 GPUs for 10 days. Nevertheless, this challenge was addressed by further research in this field by proposing efficient performance estimation strategies.

The third element of these algorithms is their performance estimation strategy. The performance estimation strategy can be seen as the method used to estimate the performance of each architecture. The computational bottleneck of most of the NAS algorithms lies in this element. In other words, evaluating each neural architecture takes a lot of time compared to other components of the proposed algorithms. Therefore, in recent years, most of the researchers have focused on this element to reduce the computational cost and increase the efficiency of their algorithms.

The simplest approach for estimating a neural network's performance is conventional training and validation of neural networks. However, since it is computationally expensive to this, recent algorithms attempted to abate this computational cost in various ways. To illustrate, one of these methods tried to solve this problem by using regression models to predict the final accuracy on the validation set of partially trained models~\citep{baker2016designing}. Additionally, other proposed algorithms applied new techniques such as early stopping~\citep{baker2017accelerating}, training on a lower subset of the actual data~\citep{Klein17}, weight inheritance (i.e., network morphism) \citep{elsken2018efficient}, and weight sharing~\citep{pham2018efficient}. However, while most of these methods effectively reduce the computational cost, they usually underestimate the real performance of the discovered architectures \citep{elsken2018neural}. By considering this, we chose to reduce the computational cost in two ways. First, we train each model for a limited but reasonable number of batches. Second, a hashing method is used to prevent the repetition of evaluating the same neural architecture.

While most of the proposed algorithms were specifically invented to work with benchmarked datasets, which are clean and well-prepared, there is a need for methods able to work with more noisy datasets that can be found in various industries, such as medicine. Thus, in this paper, our main concentration is on developing a NAS algorithm that can work well with such datasets. In more detail, to address the problems of data shortage and data imbalance, we use a data aumentation technique and an oversampling method, respectively. But, these techniques cause high fluctuations in validation accuracy during training, which makes it hard to measure when a model converges. To cope with these challenges, a special technique named GeNAS-WF as a performance estimation strategy is introduced in this paper.

\subsection{Automatic Sperm Processing}
\label{automatic_sperm_processing}

The automatic selection of sperms has been the objective of many studies. In one of the studies, \cite{ramos2004evaluation} combined computerized karyometric image analysis (CKIA)
system and DNA-specific stain (Feulgen)  for evaluation of ICSI-selected epididymal sperms. They have used a high magnification (1000$\times$) microscope.

In another research, the fraction of boar spermatozoa heads was measured and a pattern for this part was trained \citep{sanchez2006statistical,sanchez2005statistical}.  In this method, a deviation model is proposed and calculated for each sperm's head. 
After that, an optimal value is obtained for the classification of each sperm. Then, sperms tails, by utilizing morphological closing, were removed and the holes in the contours of the heads were filled. In the end, by applying Otsu's method \citep{otsu1979threshold}, the head of each sperm is separated from the background.

In \citep{vicente2013comparative}, sperm nuclear morphometric subpopulations of different species including goat, sheep, pig, and cattle were processed using ImageJ \citep{abramoff2004image} and the results were used for multivariate cluster analyses. There is also another work in which the effects of different staining methods on the human sperm head were reported \citep{maree2010morphometric}. In their study, stained and fresh sperms were compared together. ImageJ also used in another method in order to assess ram sperm morphology on the stained images \citep{yaniz2012automatic}. \citep{zhang2017animal} has also proposed a novel method for animal sperm morphology analysis. 
Different algorithms such as active contour model, K-means, thinning algorithm, and image moment have been utilized in this method.

In another research, the Bayesian classifier was applied in order to extract different parts of sperm: acrosome, nucleus, midpiece, and tail \citep{bijar2012fully}. This segmentation was done using Markov random field model and the entropy-based expectation-maximization algorithm on the stained human semen smear. The images were captured with a high-magnification (1000$\times$) microscope.

\cite{abbiramy2010spermatozoa} proposed a method for the classification of sperms into normal and abnormal classes. Their method proceeds in four steps: 1) image preprocessing: RGB to grayscale conversion and noise removal by applying median
filter; 2) sperm detection and extraction using the Sobel edge detection algorithm; 3) segmentation of each sperm; and 4) applying classification to detect normal and abnormal sperms.

In another study, Combining learning vector quantization (LVQ) and digital image processing was used for the classification of boar sperm acrosome \citep{alegre2008automatic}.
The images were captured using a phase-contrast microscope. This method works on stained
images, and the experimental results have shown a 6.8\% error on the classification task.

A combination of histogram statistical analysis and clustering techniques is another method that has been applied in sperm detection and segmentation \citep{chang2014gold}.
In another research, principal component analysis (PCA) was also applied in order to extract features from sperm images \citep{li2014human}. K-nearest neighbors (KNN) technique was also used for the classification of normal sperms. There are also some methods that focus on microscopic videos for sperms segmentation and calculation of their motilities \citep{haugen2019visem, boumaza2018automatic, ilhan2018novel}.

One of the successful methods for normal sperm selection, which is able to work with fresh human sperms, is the algorithm of \cite{ghasemian2015efficient}. This method works with images from a low-magnification microscope (400$\times$ and 600$\times$), and the images are non-stained. One of the other advantages of this method is its real-time processing time.  

To the best of our knowledge, there are only a few researchers that applied deep learning for normal sperms classification. In one of these studies, the sperms DNA integrity has been
predicted from sperm images using a deep CNN \citep{mccallum2019deep}. They have trained CNN on a collection of approximately 1,000 sperm cells of known DNA quality, for prediction of DNA quality from brightfield images. A pre-trained CCN architecture (i.e., VGG16) has been used in this study, and some additional layers were added after the last convolutional layer. The achieved results were acceptable in terms of DNA integrity prediction.

In another deep learning method, sperm images were classified into the World Health Organization (WHO) shape-based categories (i.e., Normal, Tapered, Pyriform, Small, and Amorphous)~\citep{riordon2019deep}. The authors also used VGG16 in order to avoid excessive neural network computation. They have applied their method on two freely-available sperm head datasets (HuSHeM~\citep{shaker2017human} and SCIAN~\citep{chang2017gold}).
The achieved results on sperm classification were superior to the other existing methods on these two datasets. However, this method cannot classify the different parts of each sperm.

One of the most successful deep learning algorithms in sperm classification is the work
of ~\cite{javadi2019novel}. In their method, after applying data augmentation techniques and
a sampling method, a deep neural network architecture was designed and trained. This architecture is able to detect morphological abnormalities in different parts of human sperm (i.e., acrosome, head, and vacuole). This algorithm was trained and evaluated on the MHSMA dataset, and the trained models were highly accurate. It should be noted that GeNAS is far more precise than this method, which is discussed in more detail in section \ref{results}.

\section{GeNAS}
\label{GeNAS}

In this section, we first present the overall algorithm of GeNAS, then we focus on the chromosome 
structure{,} and the fitness function. Finally, the primary customized operations of selection, crossover, 
and mutation are described.

\subsection{The overall structure of GeNAS}
\label{the_overall_flowchart_of_GeNAS}

\begin{figure}[H]
\begin{center}
\includegraphics[width=0.6\linewidth]{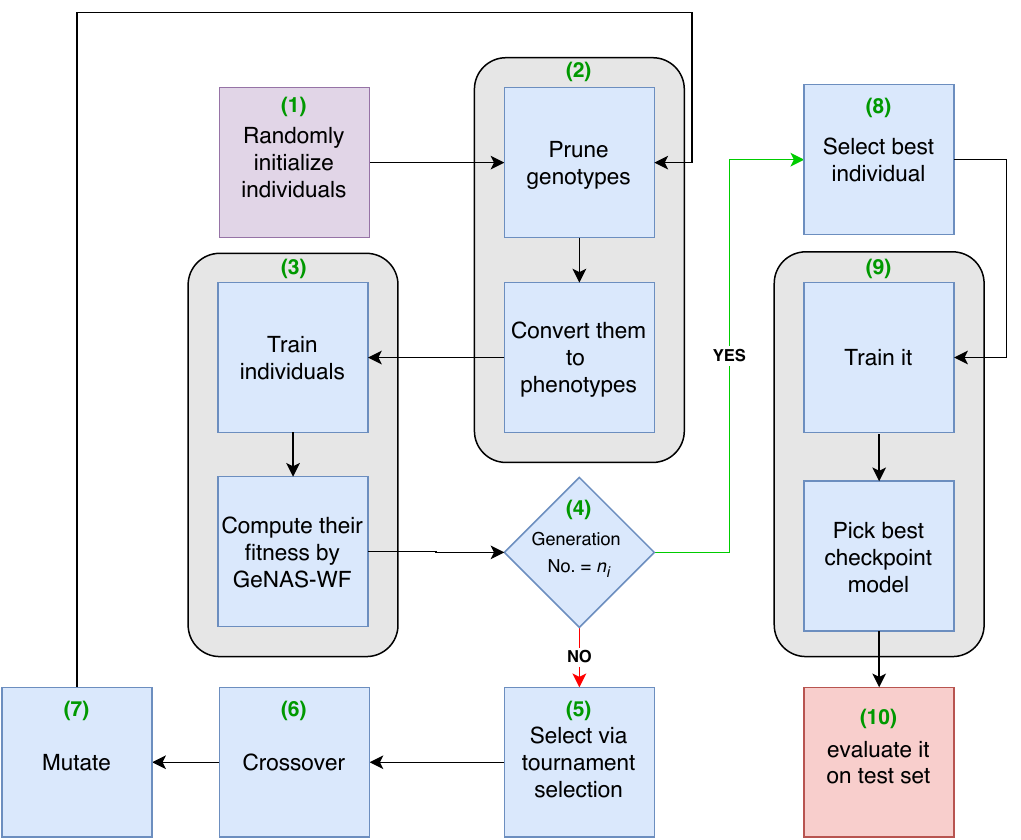}
\end{center}
\caption{Structure of GeNAS}
\label{gnas_architecture}
\end{figure}

The overall scheme of GeNAS is shown in \ref{gnas_architecture}. The algorithm starts with initializing the population by generating $n_p$ number of chromosomes. The process of generating each chromosome consists of two steps: first, we set the length of the chromosome by sampling a random value from the feasible set; then, the value of each gene of the chromosome will be selected from a constrained search space, which is described in subsection \ref{search_space}. 
Each chromosome's phenotype corresponds to a CNN architecture that consists of convolutional cells. Each convolutional cell is made of a convolution layer, followed by a max-pooling layer.

During the genotype $\to$ phenotype translation process, a pruning operation will take place if the genotype is not feasible, i.e., the output of the corresponding convolutional neural architecture has negative height and weight values. Briefly, the pruning operation will cut some of the convolutional cells in the chromosome's head to make it feasible.

Next, the phenotype of each individual is trained on $n_b$ number of mini-batches, and, after training on each mini-batch, accuracy on the validation set is evaluated and saved. These accuracies will be used, at a later stage, to compute the fitness of the individual using the GeNAS-WF technique, which is explained in subsection \ref{fitness}. Then parents selection is performed on the population using tournament selection with a tournament size of $\dfrac{n_p}{3}$. At last, a special crossover operation with a probability of $p_c$, followed by a mutation operation with a probability of $p_m$, is applied to the selected parents to produce a new population. The crossover operation can change the length of each child and helps to not only explore the search space of neural architectures with different lengths but also exploit the best individuals in the population.
On the other hand, the mutation operation will change 
genes' value in genotype, so it is responsible for exploring the different number of filters and stride-size in the phenotype.

These steps will be repeated for $n_i$ number of times. Then, the architecture with the best fitness among all populations will be chosen as the optimal neural architecture.

At the final stage, the selected architecture will be trained on a higher number of mini-batches from scratch. During this step, after training on each mini-batch, the accuracy of the model on the validation set will be evaluated, and the model with a maximum of this accuracy will be saved. At the last step, the saved model will be assessed on the test set.

The overall structure of GeNAS is summarized as follow:
  \begin{enumerate}
    \item Randomly initialize each individual of the first generation from the constrained search space.
    \item Prune the genotype of each individual and translate it to the corresponding phenotype, as shown in \ref{genotype_to_phenotype}.
    \item Train the individual for $n_b$ number of mini-batches, then use the GeNAS-WF method to compute its fitness value.
    \item If generation $n_i$ is reached, go to step 8.
    \item Perform Tournament Selection with tournament size of $\dfrac{n_p}{3}$ to select parents for crossover operation.
    \item Perform the special crossover, introduced in subsection \ref{crossover}, with the probability of \textit{$p_{c}$}.
    \item Perform mutation, explained in subsection \ref{mutation}, with probability of \textit{$p_{m}$}, then go to step 2.
    \item Select the individual with the maximum fitness value among all individuals of all populations as the optimal individual.
    \item Train the optimal individual for $n_b'$ number of mini-batches, and, during training, save the model which has the maximum accuracy on the validation set.
    \item Evaluate the optimal trained model on the test set and report the test measures on the test set.
    
  \end{enumerate}

\subsection{Chromosome Structure}
\label{chromosome}

\begin{figure}[H]
    \centering
    \includegraphics[width=0.5\linewidth]{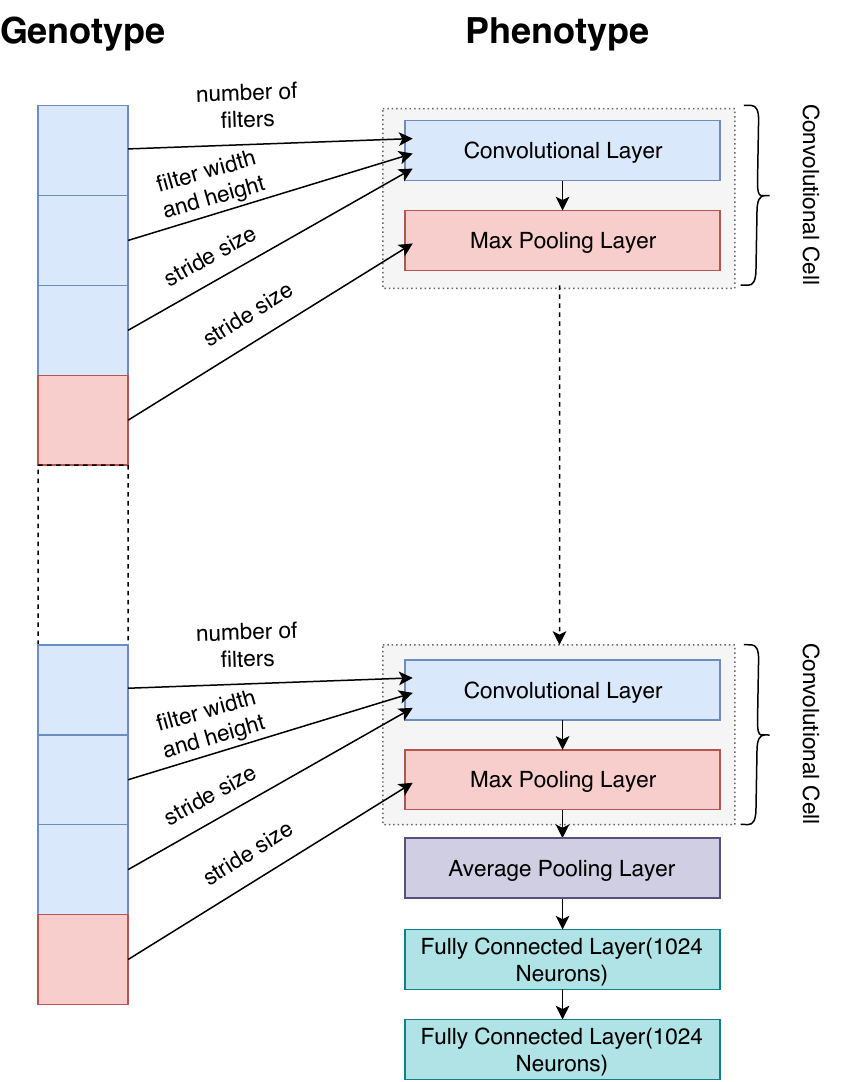}
    \caption{Translation of the genotype of a chromosome to its phenotype in GeNAS}
    \label{genotype_to_phenotype}
\end{figure}

    The proposed chromosome is a linear chromosome with discrete genes' values, which its genotype encodes the properties of the architecture of a CNN (phenotype). In this encoding scheme, each chromosome encodes features of multiple convolutional cells. A convolutional cell is composed of a convolutional layer following by a max-pooling layer. Starting from the first gene, every four distinct consecutive genes of a chromosome represent the features of a convolutional cell. Therefore, a chromosome of length $n_c$ contains $\dfrac{n_c}{4}$ number of convolutional cells, and every chromosome's length should be a factor of four. In more detail, the first three genes of a convolutional cell encode the number of filters, the filter-size (width and height of filters considered equal), and the stride-size of a convolutional layer. The fourth gene represents the stride-size (stride-size and max-pool window size are considered equal) of a max-pooling layer. Furthermore, while features of convolutional cells differ depend on the chromosome's values, the last three layers of the phenotype are consistent for all the chromosomes. The structure of each genotype and its translation to phenotype is shown in ~\ref{genotype_to_phenotype}.
    
\begin{figure}[H]
    \centering
    \includegraphics[width=0.5\linewidth]{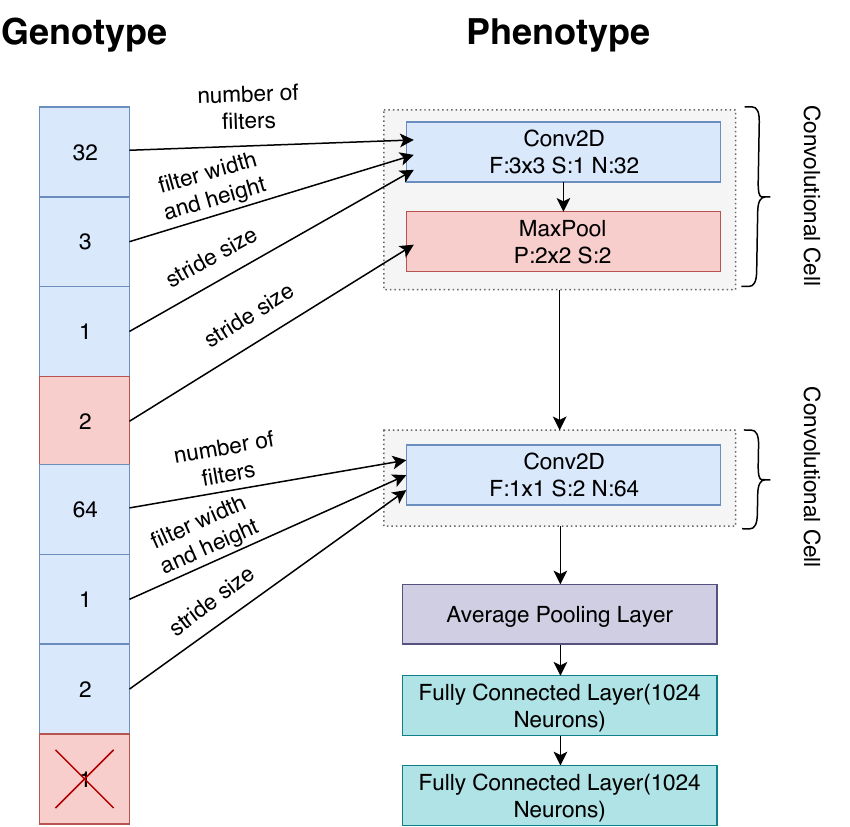}
    \caption{A detailed example of genotype translation of a chromosome to its phenotype (Note that the last max-pooling layer is removed because its stride-size is 1)}
    \label{genotype_to_phenotype_sample}
\end{figure}

    Due to the definition of the chromosome, our proposed linear chromosome can describe any plain convolutional neural architecture. To illustrate, if we want to have consecutive max-pooling layers in our convolutional neural architecture, we can allow our search space to assign 0 value to the filter height and width size, which is equal to not having a convolutional layer. Additionally, when we want to have consecutive convolutional layers in our convolutional neural architecture, we can allow our search space to assign value 1 to the stride-size (pooling-size) of a max-pooling layer, which is equal to not having a max-pooling layer. ~\ref{genotype_to_phenotype_sample} shows an example of genotype decoding.

    \subsection{Pruning}
    \label{pruning}
    
\begin{figure*}[h]
    \centering
    \includegraphics[width=.9\textwidth]{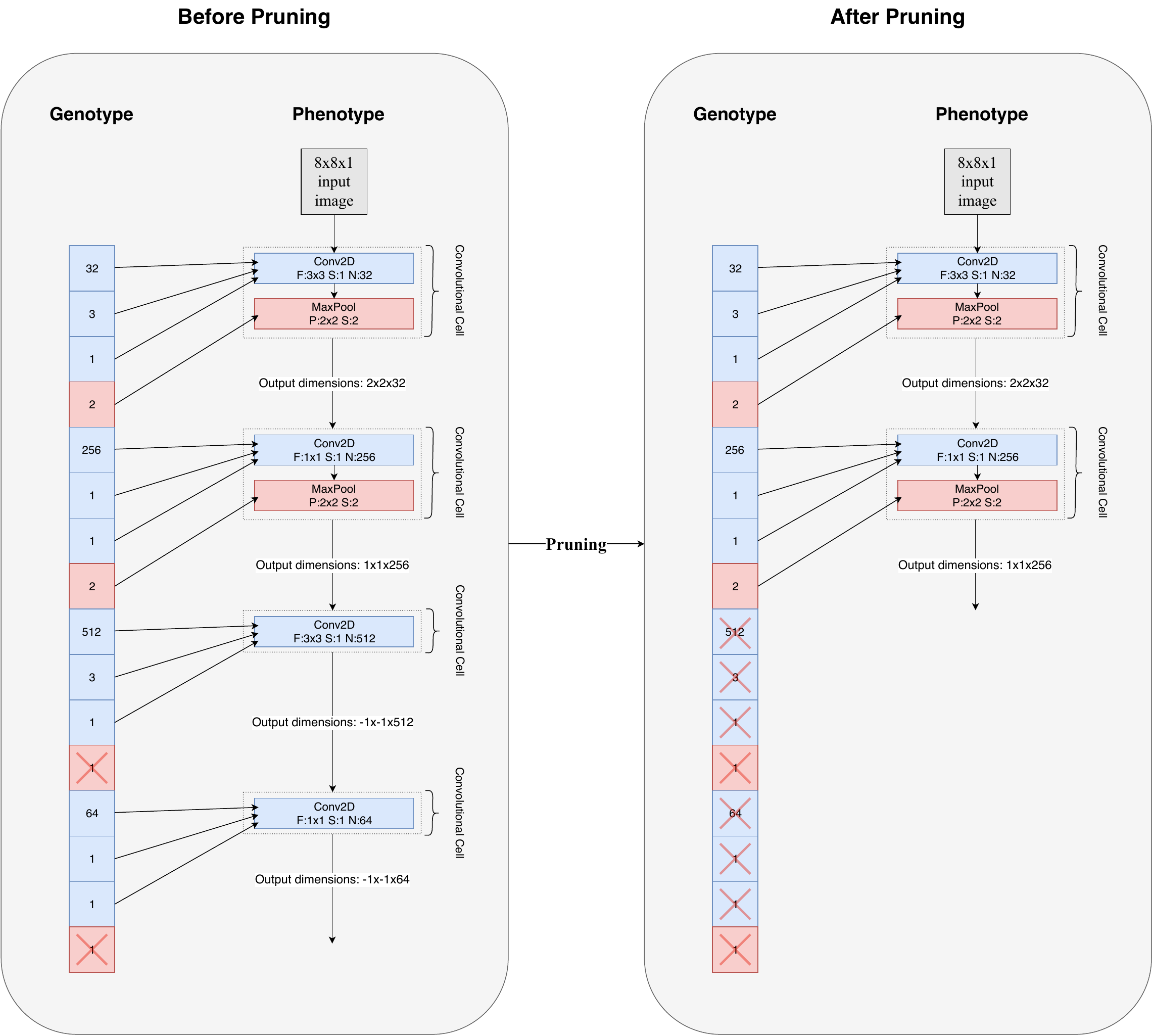}
    \caption{Pruning of a chromosome with four convolutional cells}
    \label{pruning_sample}
    \end{figure*}

    Before the fitness evaluation step, to make sure that the chromosome is feasible, a pruning process will proceed. A chromosome considered infeasible when, by stacking its corresponding convolutional cells, we get a negative output from the constructed CNN. In the process of genotype to phenotype translation, this happens because often after we add a convolutional or a max-pooling layer in each step to the phenotype, output dimensions of the phenotype will abate, according to equations \ref{eq:1}, \ref{eq:2}, \ref{eq:3}, and \ref{eq:4}. For clarification, an example of the pruning process on a chromosome with four convolutional cells is illustrated in the \ref{pruning_sample}. In this example, the padding size employed is zero for all convolutional cells, so equations \ref{eq:3} and \ref{eq:4} are only used for calculating the output dimensions.
    
    In simple terms, when the output of a phenotype gets negative, we cut enough convolutional cells from its head, until we get positive output. However, both for taking this process in parallel with the genotype to phenotype translation into account and accelerating it, we took another approach in practice. In this approach, before adding each convolutional or max-pooling layer, we calculate the output dimensions of the whole network, with the help of equations \ref{eq:1}, \ref{eq:2}, \ref{eq:3}, and \ref{eq:4}. Then, if we get a negative output, the translation process will be stopped, and the constructed CNN will be sent for fitness evaluation; otherwise, we add the layer to the top of the phenotype and repeat this process. 

    \begin{equation}    \label{eq:1}
        W_{new} = (W_{current} - F + 2P)/S + 1 
    \end{equation}
    \begin{equation} \label{eq:2}
      H_{new} = (H_{current} - F + 2P)/S + 1
    \end{equation}   
    \begin{equation}   \label{eq:3}
        W_{new} = (W_{current} - F)/S + 1
    \end{equation}
    \begin{equation} \label{eq:4}
        H_{new} = (H_{current} - F)/S + 1
    \end{equation}
    Where $W_{new}$ and $H_{new}$ are the new weight and height size, $W_{current}$ and $H_{current}$ are the current weight and height size,
    F is the filter-size, S is the stride-size, and P is the padding-size. These equations work both for calculating the dimension of
    convolutional and max-pooling layers' output. The equations \ref{eq:1} and \ref{eq:2} take place when we use padding, otherwise, the equations \ref{eq:3} and \ref{eq:4} are used.  
    
    \subsection{Fitness Evaluation}
    \label{fitness}
    
    One of the indispensable components of a genetic algorithm paradigm is formulating its fitness function. In single-objective optimization problems, a good fitness function acts as an objective function that maps a feasible solution to a scalar value, which is a summarization of its closeness to a set of desired aims. In the field of machine learning, the common scalar value which has been used to evaluate the performance of a neural network is the accuracy of it on the validation set. Hence, since we want to discover a high-performance model, and using the validation accuracy worked well for the prior studies in the NAS domain, the fitness value is obtained using the accuracy of the model on the validation set. 
    
\begin{figure}[H]
        \includegraphics[width=\linewidth]{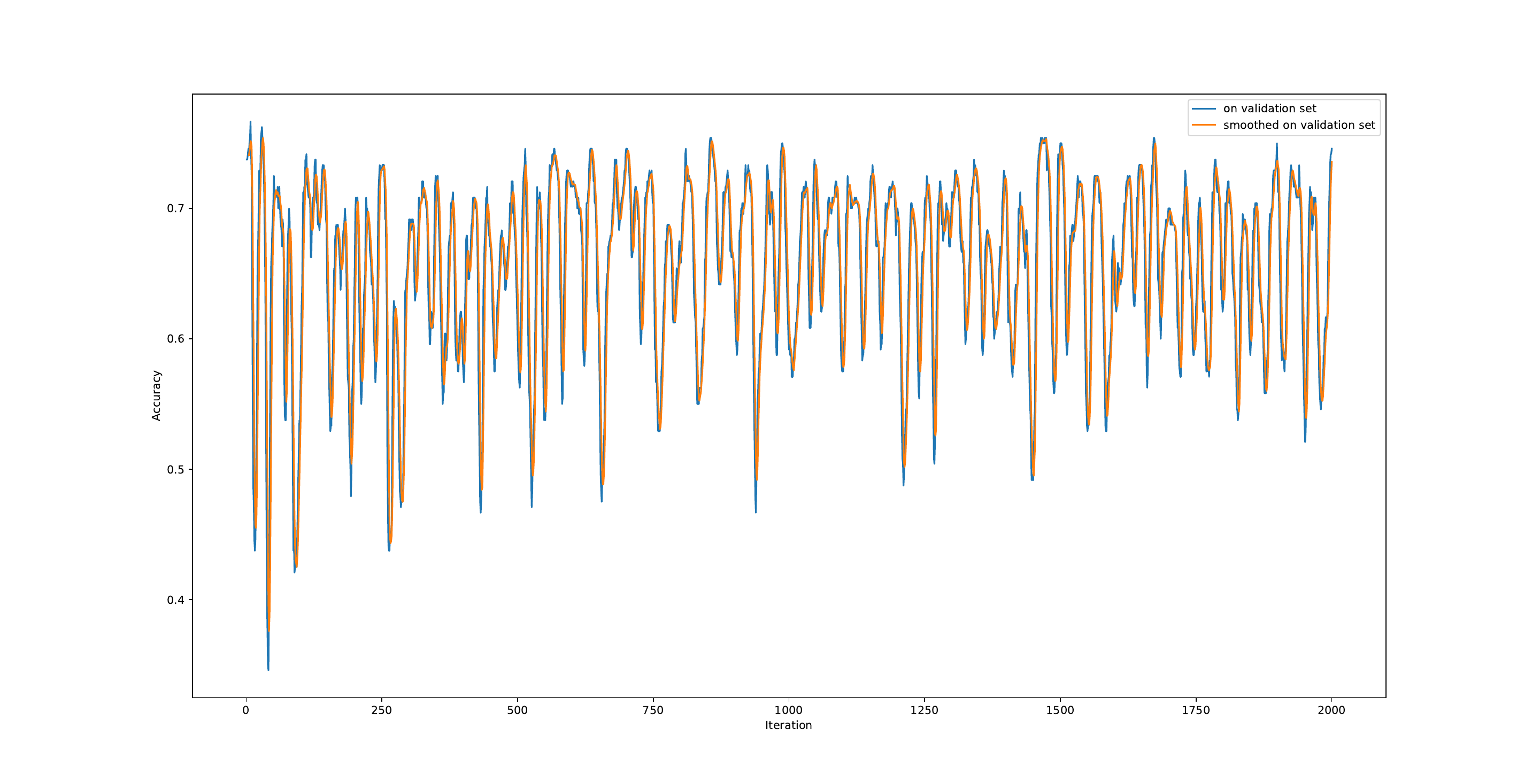}
        \caption{Fluctuations of a sample CNN accuracy on validation set through iterations on head label}
        \label{sample_model_training_plot}
    \end{figure}

    However, in comparison with previous studies, we used a different technique to gain fitness value from validation accuracy. The reason for proposing this technique is our experiments revealed that validation accuracy inordinately fluctuates during the training phase. One example of these fluctuations is shown in \ref{sample_model_training_plot}. The cause of these fluctuations lies in using a special sampling and data augmentation methods (explained in section \ref{experiments_results}) during the training phase, as sub-components of GeNAS, to cope with imbalanced data and shortage of data respectively. We termed this technique GeNAS-WF and, briefly, it helps us to level off these fluctuations.
    
    In this technique, to obtain this fitness value for each solution, first, we train it on $n_{iter}$ number of mini-batches. In the course of the training phase, after training on each mini-batch, the accuracy of the model on the validation set will be accumulated in a vector named \textit{B}. Just to clarify, the first element of vector \textit{B} carries validation accuracy after training on the first mini-batch, the second element carries validation accuracy after training on the second mini-batch, and it follows this pattern until the last element, which is $n_{iter}^{th}$ element. In the next step, we decide on a customized window, named W, which is a vector of size $n_w$. Then, the cross-correlation between \textit{B} and \textit{W} will be calculated. This operation will result in a smoother vector \textit{G}. For illustration, the first element of \textit{G} is computed by calculating the weighted mean of one to  $n_w$ elements of vector \textit{B}, the second element is computed by calculating the weighted mean of two to $n_w + 1$ elements of vector \textit{B} and so on, using \textit{W} as the weights. At last, the maximum element of vector \textit{G} will be considered as the fitness value of the model. The equations are as follows:

    \begin{equation} \label{cross-correlation}
        G[i] = \sum_{u=0}^{n_w} W[u]*B[i+u] \; \; for \; i=0,1,2,...,(n_{iter}-n_w+1) \\
    \end{equation}
    
    \begin{equation} \label{max_cc}
        fitness = \max_{i=0}^{n_{iter}-n_w+1}(G[i])
    \end{equation}
     
    In the equations above, characters in the square brackets refer to the specific element of their corresponding vectors. For example, in the \textit{G[i]} notation, the character \textit{i} indicates the $i^{th}$ element of vector \textit{G}.
    
\begin{figure}[H]
        \includegraphics[width=\linewidth]{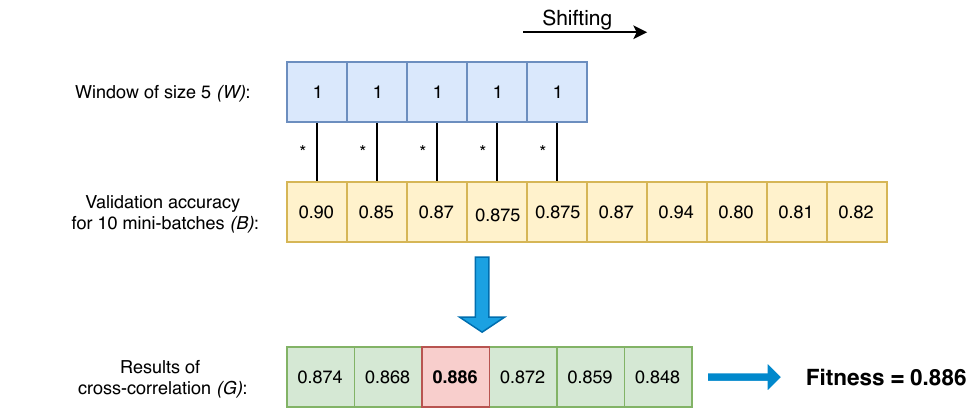}
        \caption{GeNAS-WF: An example of calculating the fitness of a trained neural architecture}
        \label{fitness_calc_example}
    \end{figure}

    Moreover, based on our experiments, we assigned the value of one to all elements of the vector $W$. Nevertheless, it has the potential to get other values for improving the performance on other datasets. An example of fitness evaluation, with a vector $W$ same as the one we used in our experiments, is illustrated in \ref{fitness_calc_example}.

% The main reason which we proposed this technique was to address the

    At last, it is good to mention that to increase the GeNAS speed and remove the redundant evaluation of individuals, before the translation process takes place, a hashing method is also proposed to check if the fitness of the genotype has already been computed, in which case, the previously computed fitness is retrieved; otherwise, the phenotype is produced, and its fitness will be computed. It is good to mention that the time-complexity of retrieving the fitness of the previously evaluated chromosome is O(1).

    \subsection{Selection}
    \label{selection}
    
    The parent selection step is performed using tournament selection with a tournament size of $\dfrac{n_p}{3}$, where $n_p$ is the population size. In this selection method, $\dfrac{n_p}{3}$ individuals are randomly chosen from the population, among which the individual with maximum fitness value is selected.

    Tournament selection has been chosen because it allows controlling the selection pressure utilizing tournament size. To clarify, as tournament size gets larger, the selection pressure gets higher and vice versa. Tournament size, therefore, will enable us to change the degree of exploration and exploitation. In more detail, by changing the tournament size, we can adapt our method to the available computational power. To illustrate, since a low amount of computational power is considered for our experiments, we selected a high value of tournament size ($\dfrac{n_p}{3}$). In this way, we increased the degree of exploitation over exploration, so our method can discover a near-optimal CNN architecture in a fair amount of time.

    \subsection{Crossover}
    \label{crossover}
    
    The crossover operation combines the genotypes of two parents to form the genotype of an offspring. The main contribution of the crossover operation of our proposed algorithm is that it helps GeNAS to change the length of the chromosomes of the new population, i.e., changing the number of convolutional cells. In our case, the crossover operation combines genotypes of different sizes and produce a genotype of yet another size.

    Crossover is applied on a pair of genotypes $parent_1$ and $parent_2$ selected through tournament selection. First, a number between zero and one is randomly chosen. If it is less than threshold $p_c$, then the crossover operation is performed on $parent_1$ and $parent_2$; otherwise, $parent_1$ and $parent_2$ are added to the new generation after a mutation step.
    
    With having in mind that the number of genes in each generated child should be a coefficient of four, the crossover is designed as follows. A crossover point $point_1$, which is an integer number between 0 and length of the parent $parent_1$,  is randomly chosen. Next, a point $point_2$ is computed for $parent_2$ in accordance to equation~\ref{eq:5}:

    \begin{equation} \label{eq:5}
	\begin{aligned}
        point_2 = RandomInteger(0, \dfrac{length(parent_2)}{4}) \times 4 + \\
						(point_1 \mod 4)
	\end{aligned}
    \end{equation}
    Where \textit{RandomInteger} function will generate a random integer between 0 and $\dfrac{length(parent_2)}{4}$ from random uniform distribution. This equation will guarantee that, in the further steps, crossover generates children which their length will be a coefficient of four.

        Once $point_1$ and $point_2$ are defined, the lengths of the two children can be computed as follows:
        
        \begin{equation} \label{eq:6}
        length(child_1) = point_1 + (length(parent_2)-point_2)
    \end{equation}
        \begin{equation} \label{eq:7}
        length(child_2) = point_2 + (length(parent_1)-point_1)
    \end{equation}
        
\begin{figure}[H]         
\includegraphics[width=\linewidth]{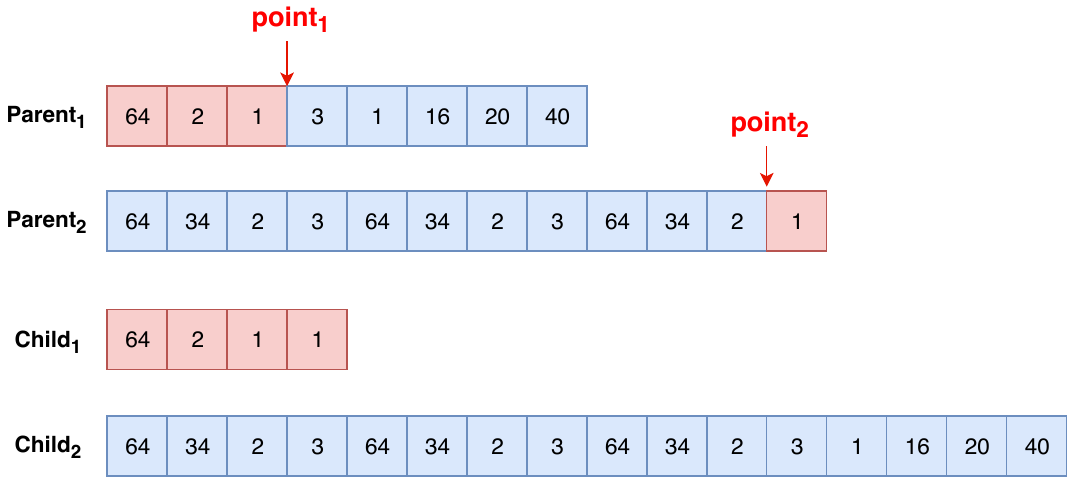}
\caption{A crossover example which generates children with different lengths.}
\label{crossover_example}
\end{figure}

After computing the children's length, if the computed lengths exceed the maximum and minimum of individuals' length in constrained search space, \textit{$point_{2}$} will be calculated again until both children's length are valid. When valid values of $point_1$ and $point_2$ have been determined, each genome is cut in two with respect to its crossover point. At the next step, the left part of $parent_1$ is concatenated to the right part of $parent_2$ to produce the first child, and the left part of $parent_2$ is concatenated to the right part of $parent_1$ to generate the second child. An example of this operation is shown in~\ref{crossover_example}.

    \subsection{Mutation}
    \label{mutation}
    
The mutation operation allows modifying the number of filters, filter-size, and stride-size of convolutional and max-pooling layers for each individual. The mutation operation will take place for each individual with mutation probability (i.e., $p_m$).
   
Considering every 4 consecutive genes represents a convolutional cell, we grouped genes into four distinct groups, each one represents a distinct feature of a convolutional cell: the number of filters, filter-size (i.e., filter width and height), convolutional stride-size, and pooling stride-size. These groups are evident in \ref{genotype_to_phenotype}. 
Considering this, when the mutation takes place for a specific chromosome, one of the genes from the selected chromosome will be picked randomly.
Later, the selected gene will be mutated differently depending on what group of genes it belongs to. If the chosen gene belongs to the number of filters or filter-size group, a random element will be selected from the respective feasible values associated with the group of the selected gene, described in section \ref{search_space}. Otherwise, if the selected gene matches with the convolutional or pooling stride-size group, a $\gamma$ value will be calculated as follows. At the first step, the current stride-size will be added to a floating-point value, which is randomly selected from a normal distribution with a mean of zero and variance of one. Subsequently, the minimum of obtained value in the previous step and respective gene group maximum constrained value, described in section \ref{search_space}, will be computed. After we obtained the $\gamma$ value, for achieving the final mutated stride-size, maximum of $\gamma$ and minimum constrained value of the respective gene group will be calculated. The equations are as follows:

\begin{equation} \label{mutation_util_formule}
	\begin{aligned}
    		\boldsymbol{\gamma} = \boldsymbol{\min(StrideSize + RandomNormal(), }\\
				\boldsymbol{MaxConstraintStrideSize)}
	\end{aligned}
\end{equation}
 
\begin{equation} \label{mutation_formule}
	\begin{aligned}
	    \boldsymbol{MutatedStrideSize} = \boldsymbol{\max(\gamma, }\\
								\boldsymbol{MinConstraintStrideSize)}
	\end{aligned}
\end{equation}
Where \textit{RandomNormal} function will generate a random value with mean 0 and variance 1 from the normal distribution. \textit{MaxConstraintStrideSize} and \textit{MinConstraintStrideSize} are the maximum and minimum values permitted to use as stride size, respectively.

\section{Experiments and Results}
\label{experiments_results}

In the following sections, the experimental part of our work is described. First, the Modified Human Sperm Morphology Analysis dataset (MHSMA), which contains annotated images of human sperm cells, is introduced. Next, the data augmentation techniques and an oversampling method, which we have designed, is illustrated. Then, constrained search space along with the modules, which we utilized in our search space, is explicated. At last, The details of GeNAS implementation and the results of GeNAS, random search, and previous benchmarks are discussed.

\subsection{Dataset}
\label{dataset}

The MHSMA dataset~\citep{javadi2019novel} is composed of 1,540 grayscale images of sperms with both size of 64x64 and 128x128. This dataset is made from Human Sperm Morphology Analysis dataset (HSMA-DS)~\citep{ghasemian2015efficient}, introduced in 2015. All images have been labeled by specialists with four binary labels: tail and neck, vacuole, head, and acrosome. The value of these labels can be either normal (positive), or abnormal (negative). The distribution of negative and positive values with respect to the four labels is given in \ref{mhsma_distro}. This table reveals that the data is highly imbalanced in favor of the positive class, which accounts for $72.86\%$ up to $95.52\%$  of the data, depending on the label.

  \begin{table}[!]
   \centering
    \caption{Data distribution of the MHSMA dataset}
    \begin{tabular}{ |l|l|l|l|l| }
      \hline
        Label & \# Positive & \# Negative & \% positive & \% Negative	 \\
      \hline
        Head  & $1,122$ & $418$ & $72.86$ & $27.14$   \\
        Acrosome & $1,086$ & $454$ & $70.52$ & $29.48$  \\
        Tail and neck & $1,471$ & $69$ & $95.52$ & $4.48$   \\
        Vacuole  & $1,301$ & $239$ & $84.48$ & $15.52$    \\
      \hline
    \end{tabular}

    \label{mhsma_distro}
    \end{table}

Furthermore, We have used the split of the data proposed by~\citep{javadi2019novel} in three subsets: training, validation, and test, which contain 1000, 240, and 300 images, respectively.

  \begin{figure}[hbt!]
	\centering
	\begin{subfigure}{0.49\linewidth}
		\centering
		\includegraphics[width=0.75\linewidth]{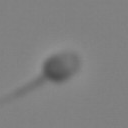}
		%\caption{a 128x128 image in MHSMA}
		\label{fig-prep-400-proc}
	\end{subfigure}
	\begin{subfigure}{0.49\linewidth}
		\centering
		\includegraphics[width=0.75\linewidth]{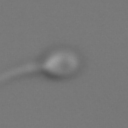}
		%\caption{b 128x128 image in MHSMA}
		\label{fig-prep-600-proc}
	\end{subfigure}
	\caption{Sample images from MHSMA dataset. As it is visibile, the images are non-stained and low-resolution.}
	\label{fig-mhsma}
    \end{figure}

     \subsection{Data augmentation}
     \label{data_augmentation}
     
In this task, each trained convolutional neural network should map an input sperm image to a 1-bit label (i.e., abnormal and normal). Most generated neural architectures by GeNAS have parameters in the order of millions. So, as the number of the neural architecture parameters increases, more training examples are needed to properly tune the parameters. But, the MHSMA dataset has only 1000 training examples, and
the act of collecting more human sperm images is extremely costly and arduous. Moreover, to the best of our knowledge, MHSMA is one of the largest datasets in the field of sperm morphology analysis. As a result, For solving the problem of training examples shortage, a data augmentation technique is adopted to prevent overfitting and virtually increasing training set size.

In this technique, before feeding each training example (sperm image) to the model, a 64x64-pixel crop will be extracted from each 128x128-pixel image. After crop extraction step, we apply random modifications to each training example which they are as follows:

   \begin{itemize}
    \item Flipping: Every image was flipped vertically and horizontally with the probability of 0.5.
    \item Rotating: The cropped area of every image was rotated by $\theta$ degrees, where $\theta$ is randomly selected regarding the uniform distribution of $[0, 360)$.
    \item Shifting: The crop region was shifted along the vertical and horizontal axis by y and x pixels, where y and x were randomly chosen regarding the uniform distribution of $[-5, 5]$.
    \item Scaling: Pixel values of each image was multiplied by $e^{\beta}$, where $\beta$ is randomly chosen regarding the uniform distribution of $[-log(1.25),log(1.25)]$.
  \end{itemize}
  
  The output of these random modifications on one training example will be a 64x64 gray-scale image. Ultimately, the image will be normalized by subtracting it from its mean and dividing the result by 255, as shown in equation~\ref{eq:8} (x is a single sperm image):
         \begin{equation} \label{eq:8}
        normal(x) = \dfrac{x - mean(x)}{255}
    \end{equation}
    It is good to mention that our augmentation settings are the same as the ones used in \citep{javadi2019novel} work.

      \subsection{Sampling}
     \label{sampling}
     As shown in \ref{mhsma_distro}, the MHSMA dataset is highly imbalanced. Hence, a special oversampling method, proposed by \cite{javadi2019novel}, is utilized to address this challenge. The main goal of this oversampling method is generating balanced mini-batches from an imbalanced dataset like the MHSMA.

    In this oversampling method, negative and positive samples will be divided into two distinct shuffled lists. The process of adding one sample to each mini-batch is as follows. First, we choose one of the lists by $0.5$ probability. Next, the first sample at the top of the chosen list will be selected, and it will be added to the mini-batch. At last, the selected sample will be moved to the end of the chosen list. For generating each mini-batch, this process will be repeated until the mini-batch fulfilled. Moreover, after all of the samples in a list is used, the list will be shuffled. By using this approach, most likely, classes in every mini-batch will be balanced.

    \begin{comment}
    \begin{table}[!]
    %%\centering
    \caption{Data Augmentation Parameters}
    \begin{tabular}{ |l|l| }
      \hline
        parameter & value	 \\
      \hline
        Shifting  & $10000$     \\
        Rotating & $0.0001$    \\
        Flipping & Vertically and horizontally    \\
        Scaling  & binary-crossentropy    \\
        Output Layer Activation & Sigmoid \\
      \hline
    \end{tabular}
    \label{neural_params}
    \end{table}
    \end{comment}

     \subsection{GeNAS Search Space}
     \label{search_space}
     Our search space contains plain convolutional architectures max-pooling and convolutional layers, with Scaled Exponential Linear Units (SELUs) \citep{klambauer2017self} as non-linearities. SELU is employed because it will keep the neuron activations close to unit variance and zero mean, so it will let us increase the depth of a convolutional architecture without considering vanishing and exploding gradients problem. 
     
     As our search space gets bigger, finding the optimal convolutional architecture will be harder and needs more time and computational power. So, with inspiration from previous popular convolutional architectures~\citep{Lecun98gradient-basedlearning,NIPS2012_4824,DBLP:journals/corr/ZeilerF13} a constrained search space is designed.

     In this constrained search space, for each convolutional layer, the meta-controller (Genetic Algorithm) should select a filter-size in \{1, 3, 5, 7, 11\}, number of filters in \{4, 8, 16, 32, 64, 128, 256\}, and convolutional cell stride-size in range of [1, 2]. Additionally, for each max-pooling layer, it should select stride-size in range of [1, 2]. In addition to these constraints, 2 up to 50 number of convolutional cells (i.e., an individual with a minimum length of 8 up to 200 genomes) is permitted. It should be noted that the same constrained search space is employed to discover a sub-optimal neural architecture for each label.
    
\begin{comment}

\begin{table}[!]
    %%\centering
    \caption{Chromosome constraints parameters}
    \begin{tabular}{ |l|l| }
      \hline
        parameter & constraints	 \\ 
      \hline
        number of filters  & $\{4,8,16,32,64,128,256\}$     \\
        filter width and height  & $\{1,3,5,7,11\}$    \\
        stride size of convolutional layer & $[1,2]$     \\
        stride size of max-pooling layer  & $[1,2]$    \\
        minimum and maximum convolutional cells  & $2$ up to $50$ \\
      \hline
    \end{tabular}
    \label{search_constraints}
    \end{table}
\end{comment}

\subsection{GeNAS Implementation Details}
\label{implementation_details}

In our experiments, the initial population consists of 30 individuals, which each of them is generated with both random length and random genes' value from the constrained search space. Moreover, the number of generations (iterations) has set to 20, and, in each of these generations, the population size will be equal to the initial population size.

Experimental details of genotype to phenotype transformation are as follows. After the input layer and convolutional cells created, one average-pooling layer with a stride-size of 2 followed by two fully connected layers with 1024 neurons will be added to the end of the phenotype. Then, one neuron with sigmoid as its activation function (output layer) will be appended to the top of the last added fully connected layer. It should be considered that the SELU is adopted as the activation function of the two fully connected layers. An example of these configurations is depicted in \ref{genotype_to_phenotype_sample}. Moreover, the SELU activation function is an Exponential Linear Unit (ELU) activation function which is scaled so that the variance and mean of the inputs will be maintained in their original state between two consecutive layers. The equations for both SELU and ELU are as follows:
    
    \begin{equation}
      selu(x) = scale \times elu(x, \alpha)
      \label{selu-equation}
    \end{equation}

    \begin{equation}
      elu(x, \alpha) =
      \begin{cases}
        x,& \text{if } x\geq 0\\
        \alpha (e^{x} - 1),& \text{otherwise}
      \end{cases}
      \label{elu-equation}
    \end{equation}
    
After the whole CNN architecture of the respective genotype is created, its weights will be initialized using the LeCun normal initializer~\citep{LeCun2012}, and its biases will be initialized to zero. In the LeCun normal initializer, the samples will be drawn from a truncated normal distribution with a standard deviation of $\sqrt{\dfrac{1}{n_{input}}}$ and zero as the center, where $n_{input}$ is the number of inputs to the weight matrix. At the next step, it will be trained for 2000 mini-batches. For optimization, ADAM optimizer~\citep{kingma2014adam}- with a constant learning-rate of $10^{-4}$, exponential decay rates for the moment estimates of $\beta_{2} = 0.999$, and $\beta_{1} = 0.9$- is employed. In detail, ADAM is an algorithm for first-order gradient-based optimization of stochastic objective functions, rooted in adaptive estimates of lower-order moments. For loss function, binary cross-entropy is applied, and each mini-batch contains $32$ training images. 

After we train the CNN, the fitness of it will be calculated by the GeNAS-WF method. For this method, we chose a window with a size of 5 and value of $[1, 1, 1, 1, 1]$, same as the window shown in the \ref{fitness_calc_example}. Concerning the tournament selection, a tournament-size of 10 is chosen that, in a later stage, will select 30 favorable parents for the next generation. Regarding the crossover and mutation operations, after experimenting with various crossover and mutation probabilities for each task, the best ones, we came up to, were 0.7 ($p_c$) and 0.3 ($p_m$) respectively. Moreover, the same algorithm, with the same settings, is used to do all three tasks which are obtaining the optimal convolutional architectures that can predict abnormality in the sperm head, vacuole, and acrosome on the MHSMA dataset.

Our experiments were done, using Keras~\citep{chollet2015keras} with Tensorflow~\citep{abadi2016tensorflow} backend on one Nvidia GPU.

\begin{comment}

\begin{table}[!]
%%\centering
\caption{Genetic Algorithm hyperparameters}
\begin{tabular}{ |l|l| }
  \hline
    parameter & value	 \\ 
  \hline
    $p_{m}$  & $0.3$     \\
    $p_{c}$  & $0.7$    \\
    Selection Method & Tournament Selection  \\
    Population  & $30$ \\
    Generation & $20$ \\
    Fitness Measure & Maximum Average over Accuracy On Validation Set of 5 consecutive mini-batches\\
    Population Initialization & Random Length and Genes' Value from Uniform Distribution \\
  \hline
\end{tabular}
\label{genetic_params}
\end{table}
\end{comment}

\begin{comment}
    
\begin{table}[!]
%%\centering
\caption{Neural Networks hyperparameters}
\begin{tabular}{ |l|l| }
  \hline
    hyperparameters & value	 \\
  \hline
    Iterations  & $2000$     \\
    Learning Rate & $0.0001$    \\
    Optimizer & ADAM    \\
    Loss  & binary-crossentropy    \\
    Output Layer Activation & Sigmoid \\
  \hline
\end{tabular}
\label{neural_params}
\end{table}

\end{comment}

\subsection{Results}
\label{results}

Before discussing our results, we want to address the measurements employed to evaluate the models found by GeNAS. In our experiments, we took accuracy, recall, precision, and $f_{0.5}$ score into account as metrics for our evaluations. The formulations of these evaluation metrics are shown in equation \ref{eq-accuracy}-\ref{eq-f-beta}, where FN, FP, TN, and TP indicate false negative (i.e., regular sperms predicted wrongly), false positive (i.e., irregular sperms predicted wrongly), true negative (i.e., irregular sperms predicted correctly), and true positive (i.e., regular sperms predicted correctly) respectively. It should be mentioned that in our experiments, for two reasons, we considered the value of $\beta$ in equation \ref{eq-f-beta} equal to $0.5$ (i.e., $f_{0.5}$). First,  previous papers practiced this same metric for their evaluations. Second, $f_{0.5}$ measure is more dependent on the precision rather than recall. Hence, since, in sperm morphology, precision is more important than recall, this value is a good metric to be used.

\begin{equation}
  \text{Accuracy} = \frac{TP + TN}{TP + TN + FP + FN}
  \label{eq-accuracy}
\end{equation}

\begin{equation}
  \text{Precision} = \frac{TP}{TP + FP}
  \label{eq-precision}
\end{equation}

\begin{equation}
  \text{Recall} = \frac{TP}{TP + FN}
  \label{eq-recall}
\end{equation}

\begin{equation}
  F_{\beta}\text{ score} = (1 + \beta^{2}) \times \frac{\text{Precision} \times \text{Recall}}{\beta^{2} \times \text{Precision} + \text{Recall}}
  \label{eq-f-beta}
\end{equation}

\begin{figure*}[htbp]
    \begin{tabular}{l r}
    \multicolumn{2}{c}{} \\
      \includegraphics[width=0.49\linewidth]{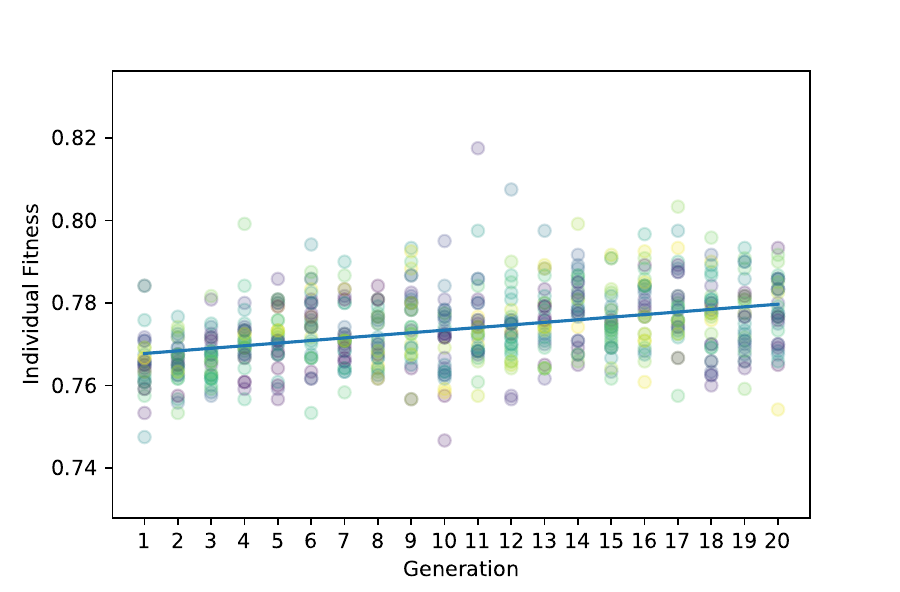} &  \includegraphics[width=0.49\linewidth]{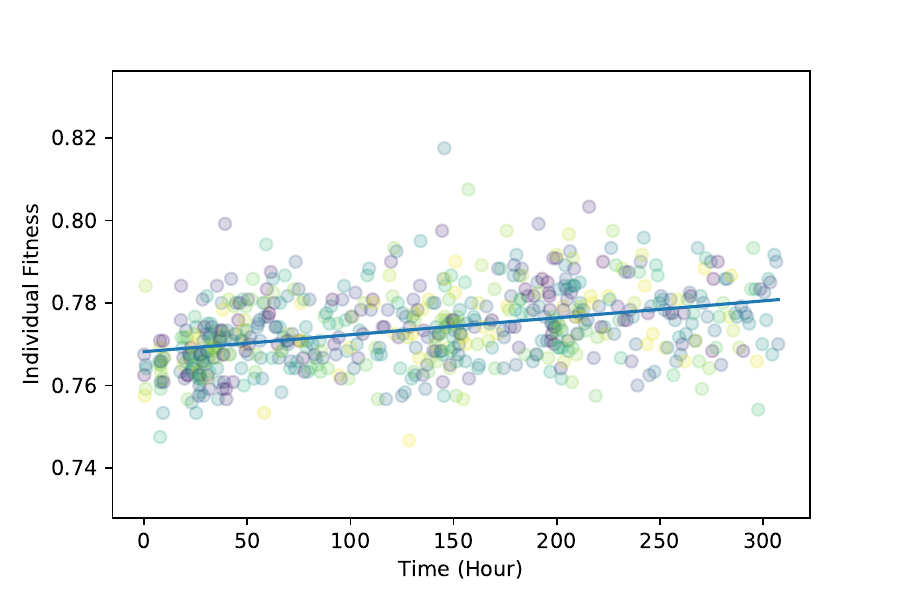} \\
    \multicolumn{2}{c}{(a) Head} \\
     \includegraphics[width=0.49\linewidth]{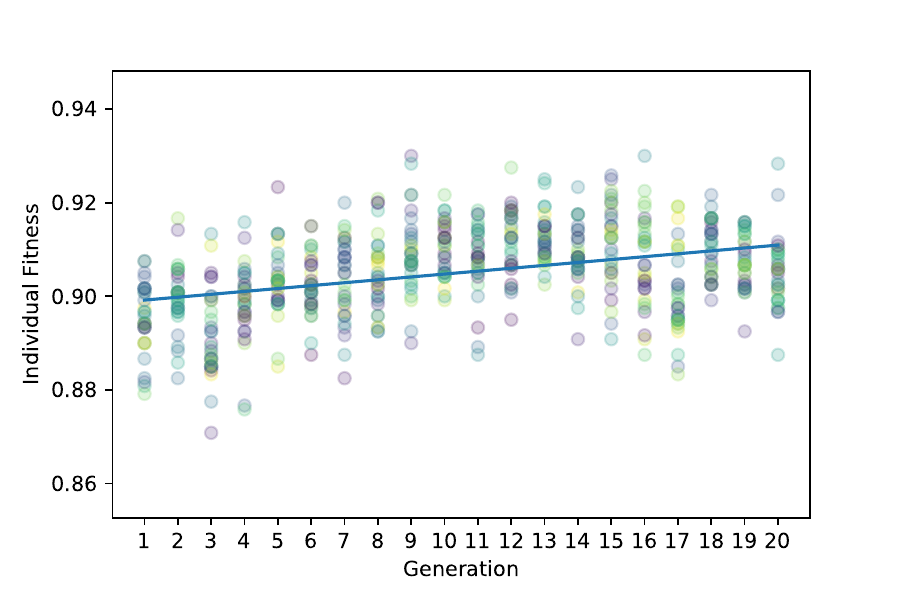} &   \includegraphics[width=0.49\linewidth]{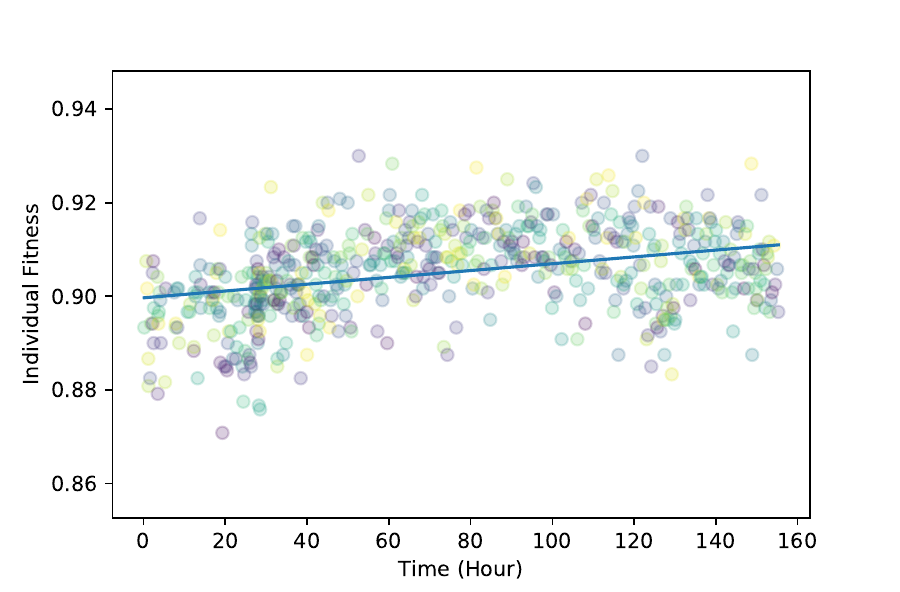} \\
    \multicolumn{2}{c}{(b) Vacuole} \\
     \includegraphics[width=0.49\linewidth]{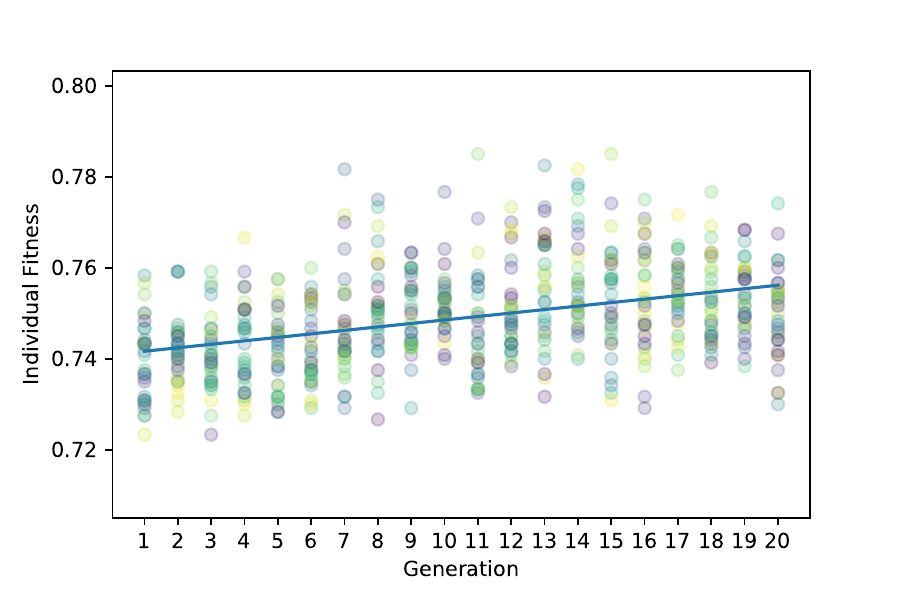} &   \includegraphics[width=0.49\linewidth]{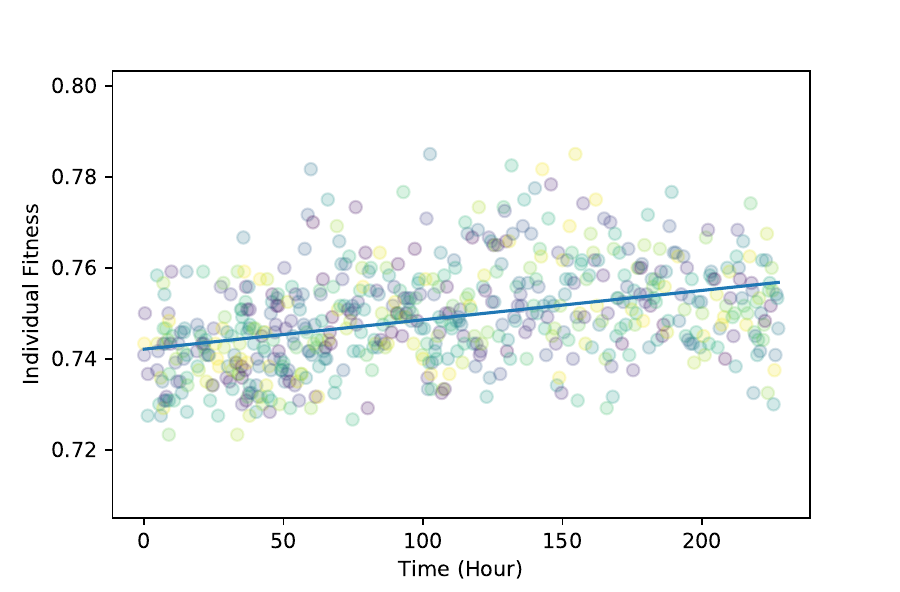} \\
	\multicolumn{2}{c}{(c) Acrosome} \\
    \end{tabular}
    \caption{Fitness of individuals through each generation and time, on the head, vacuole, and acrosome label, vertically (Colors are just for clarification and they do not represent any value)}
    \label{fitness_time_charts}
    \end{figure*}

We have employed GeNAS to discover the best architectures, which can predict abnormality in the sperm head, vacuole, and acrosome independently. After the meta-controller trained and evaluated on 600 architectures (i.e., 30 individuals in 20 generations), we extracted the architecture with the best fitness. The searching process for each label shown in the \ref{fitness_time_charts}. According to this figure, for each label, the overall fitness has gradually increased during the evolution process. For clarification, the linear regressions of all the points in each chart in \ref{fitness_time_charts} are calculated and shown. With respect to these charts, GeNAS took approximately $310$, $162$, and $240$ hours on just one GPU to run for the head, vacuole, and acrosome respectively. Therefore, considering that designing these architectures will take plenty of time for a human expert, it is less time consuming and less arduous to use GeNAS instead of hand designing these architectures.
      
\begin{figure}[htbp]
  \centering
  \begin{subfigure}{0.49\textwidth}
    \centering
  
    \includegraphics[width=1\linewidth]{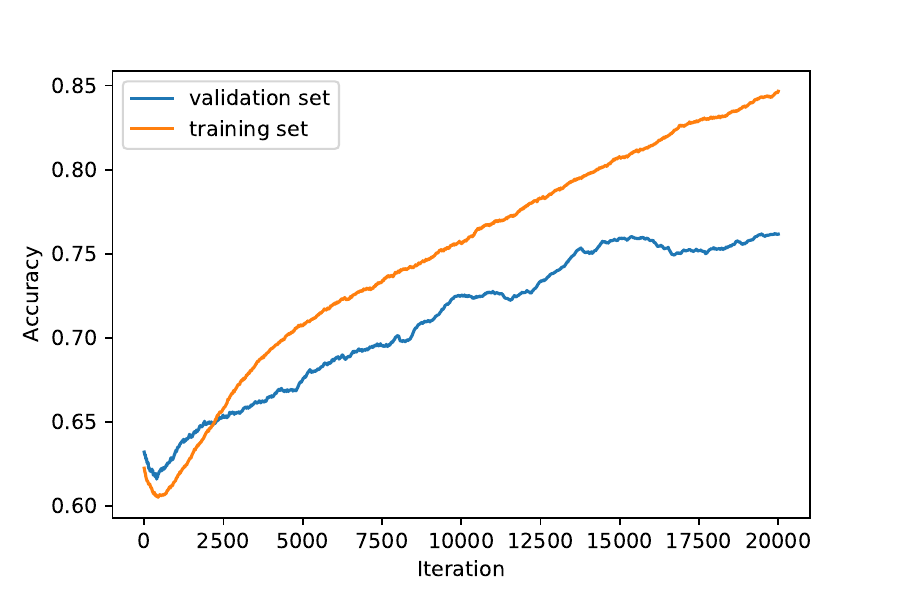}
      \caption{Acrosome}
    % \label{fig-acc-acrosome}
  \end{subfigure}
  \begin{subfigure}{0.49\textwidth}
    \centering
    \includegraphics[width=1\linewidth]{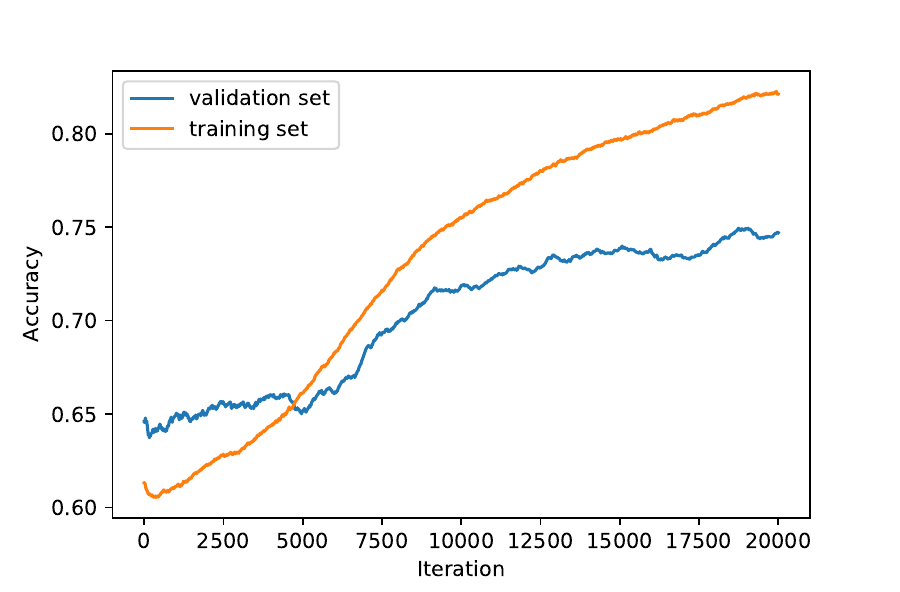}
        \caption{Head}

    % \label{fig-acc-head}
  \end{subfigure}
  \begin{subfigure}{0.49\textwidth}
    \centering
    \includegraphics[width=1\linewidth]{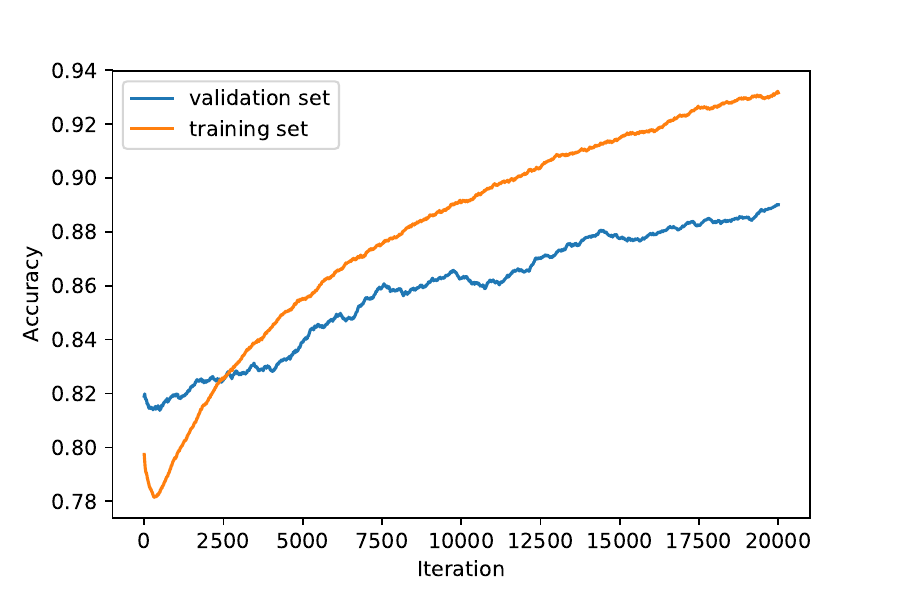}
        \caption{Vacuole}

    % \label{fig-acc-vacuole}
  \end{subfigure}
  \caption{Training and validation accuracy over iterations of training}
  \label{iteration_accuracy}
\end{figure}

After the best chromosome extracted, it has been trained for 20,000 mini-batches. The validation accuracies, for each label over iterations of training, are illustrated in \ref{iteration_accuracy}. It should be noted that, in this figure, curves have been smoothed using a technique termed Simple Exponential Smoothing~\citep{gardner1985exponential}. 

\begin{figure}[H]
\centering
\includegraphics[width=.95\linewidth]{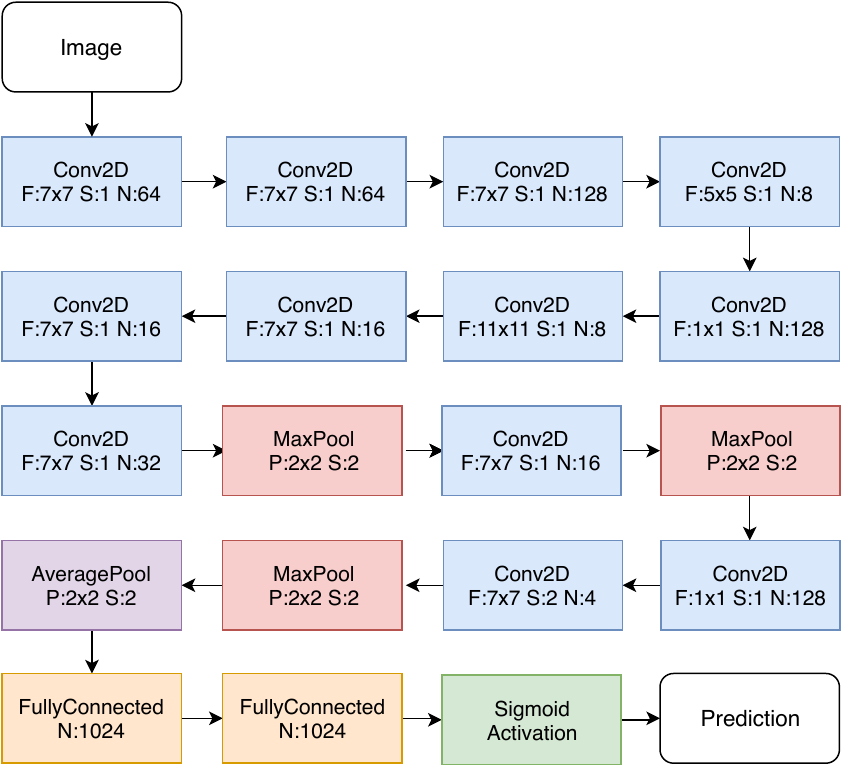}
\caption{Architecture of best model for sperm head label}
\label{best_head_cnn}
\end{figure}

\begin{figure}[H]
\includegraphics[width=.95\linewidth]{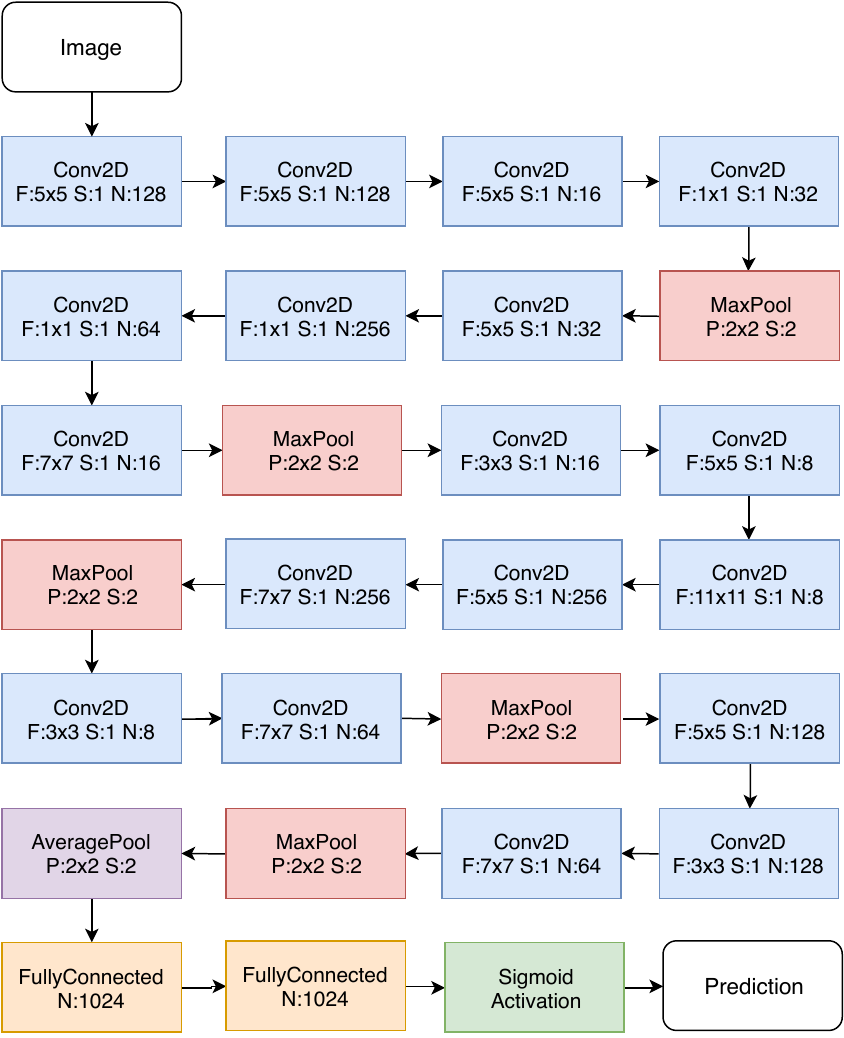}
\caption{Architecture of best model for sperm acrosome label}
\label{best_acrosome_cnn}
\end{figure}

\begin{figure}[H]
\includegraphics[width=.95\linewidth]{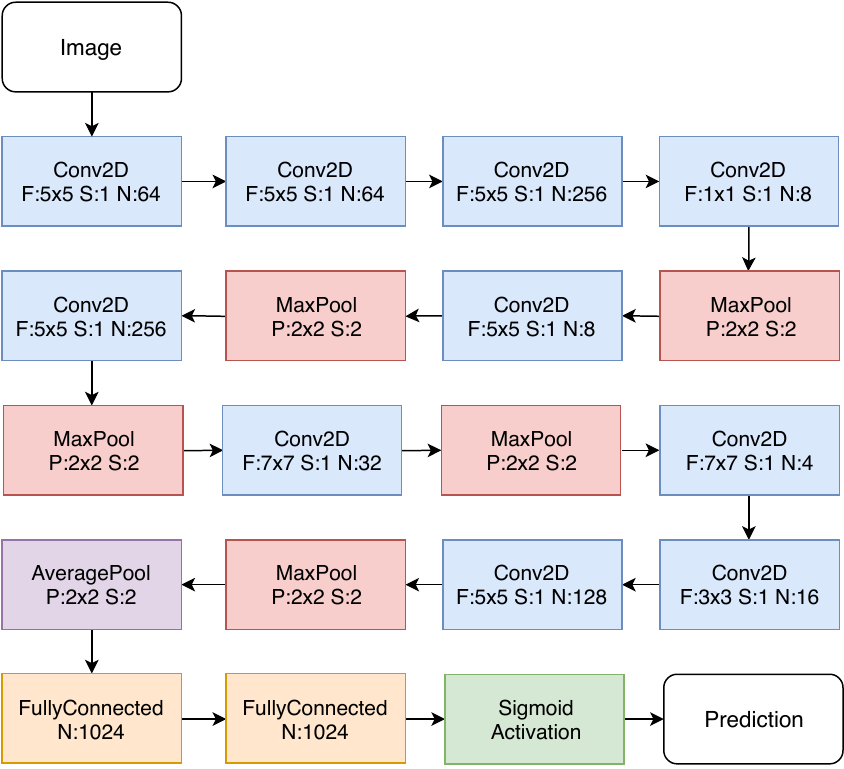}
\caption{Architecture of best model for sperm vacuole label}
\label{best_vacuole_cnn}
\end{figure}

The best discovered architectures are shown in \ref{best_head_cnn},  \ref{best_acrosome_cnn}, and  \ref{best_vacuole_cnn}.
For the head label, illustrated in \ref{best_head_cnn}, the neural architecture found by GeNAS consists of 12 convolutional and three max-pooling layers.

For the acrosome label, represented in \ref{best_acrosome_cnn}, the neural architecture discovered by GeNAS includes 18 convolutional and five max-pooling layers.

For the vacuole label, shown in \ref{best_vacuole_cnn}, the neural architecture ascertained by GeNAS is formed of 10 convolutional and five max-pooling layers.

At last, the discovered neural architectures were evaluated on the test set. The confusion matrix was generated and represented in \ref{table-confusion}.

\begin{table}[htbp]
  \centering
  \caption{Confusion matrix for evaluation of best models found by GeNAS on test set, for each label}
  \label{table-confusion}
  \begin{tabular}{|l|l|l|l|}
    \hline
    \multirow{2}{*}{Label}    & \multicolumn{2}{c}{Actual class}        & \multirow{2}{*}{Predicted class} \\
                              & Normal             & Abnormal           &                                  \\
    \hline
    \multirow{2}{*}{Acrosome} & 178 true positives & 32 false positives & Normal                           \\
                              & 35 false negatives & 55 true negatives  & Abnormal                         \\
    \hline
    \multirow{2}{*}{Head}     & 185 true positives & 34 false positives & Normal                           \\
                              & 34 false negatives & 47 true negatives  & Abnormal                         \\
    \hline
    \multirow{2}{*}{Vacuole}  & 249 true positives & 12 false positives & Normal                           \\
                              & 13 false negatives & 26 true negatives  & Abnormal                         \\
    \hline
  \end{tabular}
\end{table}

To the best of our knowledge, the only paper designed a convolutional neural network for the MHSMA dataset is \citep{javadi2019novel} paper. As shown in \ref{table-layers}, in this paper, the same neural architecture composed of two max-pooling and 24 convolutional layers followed by one average-pooling layer and two fully-connected layers is proposed for predicting the abnormality on each label. On the other hand, GeNAS identified distinct neural architectures for each label. Additionally, in comparison with \cite{javadi2019novel} paper, the best architecture found by GeNAS consists of less than half the number of convolutional layers and three more max-pooling layers for the vacuole label; half the number of convolutional layers and one more max-pooling layers for the head label; and 6 less number of convolutional layers and three more max-pooling layers for the acrosome label. At last, it is good to mention that the same number of average-pooling and fully-connected layers with the same number of neurons are used in all architectures. The results of this method are shown as ``Javadi et al." in the \ref{comparison_methods}.

\begin{table*}[htbp]
  \centering
  \caption{Comparison of number of different layers in GeNAS and Javadi et al. paper \citep{javadi2019novel}}
  \label{table-layers}
  \begin{tabular}{|l|l|l|l|l|l|l|}
    \hline
    \multirow{1}{*}{Label} & method    & convolutional & max-pooling & fully-connected & average-pooling  & total                           \\
    \hline
    \multirow{2}{*}{Acrosome} & GeNAS & \textbf{18} &  5 & 2 & 1 & \textbf{26}       \\
                              & Javadi et al. & 24 & \textbf{2} & 2 & 1  & 29                     \\
    \hline
    \multirow{2}{*}{Head}     & GeNAS & \textbf{12} & 3 & 2 & 1 & \textbf{18}               \\
                              & Javadi et al. & 24  &  \textbf{2} & 2 & 1  & 29                      \\
    \hline
    \multirow{2}{*}{Vacuole}  & GeNAS & \textbf{10} & 5 & 2 & 1  & \textbf{18}                      \\
                              & Javadi et al. & 24  & \textbf{2} & 2 & 1  & 29                     \\
    \hline
  \end{tabular}
\end{table*}

Another experiment that we conducted was about running a random search to find the best architectures which can predict abnormality in the sperm head, vacuole, and acrosome. In the beginning, random search trained 600 distinct CNN architectures (similar to our proposed method), randomly generated from constrained search space. Then, every trained architecture was evaluated on the validation set, and the architecture with the highest validation accuracy selected. In the next step, the selected architecture was trained for 20,000 mini-batches (i.e., the same as GeNAS). After training on each mini-batch, accuracy on the validation set computed and the checkpoint with the highest validation accuracy saved. Finally, the trained model evaluated on the test set and test accuracy, precision, recall, and $f_{0.5}$ score calculated. The results are shown in the \ref{comparison_methods}.

Another paper proposed an image processing based algorithm to predict abnormality on the human sperm morphology analysis dataset (HSMA-DS), which is the dataset that MHSMA based on \cite{ghasemian2015efficient}. Their algorithm has been assessed on two of the labels: head and vacuole. The results of this method are shown as ``Ghasemian et al." in the \ref{comparison_methods}.

\begin{table*}[htbp]
  \centering
  \caption{Comparison of results of best models found by GeNAS with other proposed methods on test set (all values are in percent except parameters)}
  \label{comparison_methods}
  \begin{tabular}{|l|l|l|l|l|l|l|}
    \hline
    Label                     & Method         & Accuracy   & Precision  & Recall  & $F_{0.5}$ score & $Parameters$  \\
    \hline
%    \noalign{\smallskip}\hline\noalign{\smallskip}
     \multirow{4}{*}{Acrosome} & GeNAS     & \textbf{77.66}      & {84.76}      & {83.56}      & {84.52}  & 5,756,553    \\
                            %   & GeNAS      & \textbf{77.66}      & 83.56      & \textbf{84.76}      & 83.80  & 5,756,553    \\
                               & Random Search      & 69.66    & 74.5      & \textbf{86.8}      & 76.67   & \textbf{1,185,209}   \\
                               & Javadi et al. & 76.67         & \textbf{85.93}      & 80.28      & \textbf{84.74}  & 5,637,649    \\
                               & Ghasemian et al.      & N/A      & N/A      & N/A      & N/A  & N/A               \\
    \hline
     %                         & G. et al.      & N/A        & N/A        & N/A        & N/A                  \\
     
%    \noalign{\smallskip}\hline\noalign{\smallskip}
                              & GeNAS     & \textbf{77.33}      & \textbf{84.47}      & 84.47      & \textbf{84.47}  & \textbf{1,908,261}    \\
                            %   & GeNAS      & \textbf{77.33}      & 84.47      & \textbf{84.47}      & 84.47   & \textbf{1,908,261}    \\
                              & Random Search      & 76.00    &  80.49     & \textbf{88.58}      & 81.98   & 3,032,401    \\
                              & Javadi et al.    & 77.00    & 83.48         & {85.39}     & 83.86   & 5,637,649      \\
      \multirow{-4}{*}{Head}  & Ghasemian et al.     & 61.00      & 76.71      & 71.79      & 75.68  & N/A               \\
     \hline
%    \noalign{\smallskip}\hline\noalign{\smallskip}
                              & GeNAS     & \textbf{91.66}      & \textbf{95.40}      & 95.03      & \textbf{95.32}  & \textbf{2,211,461}    \\
                            %   & GeNAS      & \textbf{91.66}     & \textbf{95.03}      & 95.40      & \textbf{95.11}   & \textbf{2,211,461}    \\
                              & Random Search      & 89.00    & 94.20      & 93.12      & 93.98    & 4,715,861     \\
                              & Javadi et al.      & 91.33    & 94.36      & \textbf{95.80}      & 94.65   & 5,637,649    \\
    \multirow{-4}{*}{Vacuole} & Ghasemian et al.      & 80.33      & 83.21      & 93.56      & 85.09  & N/A             \\

    % \noalign{\smallskip}\hline\noalign{\smallskip}
    % \multirow{2}{*}{Tail}     & Ours           & N/A        & N/A        & N/A        & N/A             & N/A        & N/A        & N/A          \\
    %                           & \bf{G. et al.} & \bf{92.33} & \bf{96.48} & \bf{95.47} & \bf{96.28}      & \bf{95.97} & \bf{59.27} & \bf{+0.1681} \\
    \hline
  \end{tabular}
\end{table*}

Eventually, as shown in \ref{comparison_methods}, GeNAS discovered convolutional architectures with higher accuracy, precision, and $f_{0.5}$ on the test set for all three labels, except the precision on acrosome. It also achieved better recall on the head label in comparison with random search and previous methods. Additionally, the discovered models by GeNAS for both the vacuole and head labels have extremely fewer parameters in comparison with other methods, with the exception of Random search on the acrosome label. However, the accuracy, precision, and $f_{0.5}$ score of the random search on this label can not compete with the value of these metrics for GeNAS. Lastly, it should be considered that critical measures for this dataset are precision and accuracy. Accordingly, the more inferior value of recall on the acrosome and vacuole labels will not harm the performance of our models.

Based on our experiments, the inference-time of all discovered models is less than 1 second, which is a proper time for treatment purposes.

\subsection{Visual explanation}
\label{sec:visual-explanation}

\begin{figure}[htbp]
    \begin{tabular}{c c c}
  
      \includegraphics[width=0.3\linewidth]{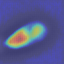} &   \includegraphics[width=0.3\linewidth]{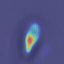}  & \includegraphics[width=0.3\linewidth]{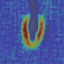}\\
      {(a) Head} & {(b) Vacuole} & {(c) Acrosome}\\

    \end{tabular}
    \caption{Grad-CAM visual explanations for each label (warmer colors illustrate more attention)}
    \label{fig-cam}
    \end{figure}

\begin{figure}[htbp]
  \centering
  \includegraphics[width=0.95\linewidth]{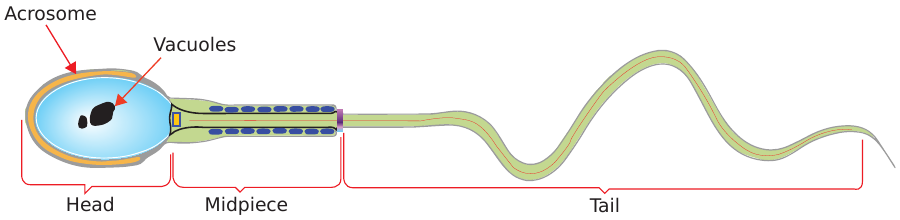}
  \caption{Diagram of distinct parts of a human sperm cell ~\citep{javadi2019novel}.}
  \label{fig-morphology}
\end{figure}

To make sure that the best neural architectures found by GeNAS pay attention to the relevant parts of sperm images for making predictions, we applied a visual explanation technique named Gradient-weighted Class Activation Mapping (Grad-CAM) \citep{kotikalapudi2017keras, selvaraju2016grad}. This visualization technique will employ class-specific gradient information to generate a heatmap, which emphasizes the areas of the input image that are essential for classification. This visual explanation will provide us with a better understanding of the function of our neural architectures. Visual explanations of models discovered by GeNAS are shown in \ref{fig-cam} for 3 different sample images which were classified accurately. These visual explanations illustrate the discovered models have certainly learned to consider the sperm image areas which are, in fact, important for the sperm abnormality prediction task. For clarification, if we compare the visual explanation shown in \ref{fig-cam} with the diagram of a sperm represented in \ref{fig-morphology}, we will be noticed that our models pay attention to the exact relevant fragments of sperm image for each classification task, i.e., parts of the head, acrosome, and vacuole of sperms.

\subsection{Discussion}
\label{discussion}

Not only we achieved better test accuracy on three distinct labels of the MHSMA dataset but also reached an outstanding precision
in comparison with prior proposed models by \cite{ghasemian2015efficient}, \cite{javadi2019novel}, and Random Search. Furthermore, GeNAS found neural architectures with extremely fewer parameters for vacuole and head labels and approximately the same number of parameters on acrosome compared with the neural architecture introduced in \citep{javadi2019novel} paper. In contrast, discovered models have reached less recall value on the head and vacuole labels compared with previous papers. But, it should be considered that precision and test accuracy are remarkably more important than recall on this dataset.

In the domain of sperm morphology analysis, especially in abnormality detection of different segments of sperms, precision and accuracy are more significant compared with other measurements because by finding even one normal sperm and using it in intracytoplasmic sperm injection (ISCI) process, one can say that treatment process can be completed successfully. Hence, since discovered models gained exceptional precision and accuracy values on all three labels, they can be used to take sperm morphology analysis a step further to real-world applications in the ISCI process. Furthermore, our discovered models can be utilized to work with cheaper medical tools like microscopes which are able to only take low-quality images mainly because our technique is specially designed to work with such images. Therefore, they can classify sperms even in poor areas that do not have high-quality tools. Lastly, it is good to consider imbalanced datasets are mainly found in the medical industry~\citep{esteva2019guide}, and GeNAS can be used to tackle many of these problems in this field in the future.

% In the light of known literature, there is no other neural architecture search algorithm designed and benchmarked on 
% imbalanced datasets, specifically for sperm morphology analysis. Broadly, to the best of our knowledge, this is the 
% first paper which introduces neural architecture search to the field of medical imaging~\citep{medical_image_analysis_survey}. 
% Therefore, in future GeNAS and new neural architecture search algorithms can be utilized to tackle various problems in 
% the medical imaging domain such as: breast cancer diagnosis~\citep{bayrak2019comparison}; 
% tissue classification of interstitial lung diseases~\citep{bondfale2018convolutional}; classification and detection of tumor cells~\citep{hossain2018detection}.

GeNAS can be employed to tackle any binary and multi-class image classification problems. It, nevertheless, would be better to customize the constrained search space of GeNAS concerning computational power and dataset characteristics. In detail, for datasets with high-resolution images, it is recommended to expand the range of stride-size both for convolutional and max-pooling layers. On the other hand, for low-resolution images, the opposite is the case. Moreover, depending on the available computational power, we can change the tournament size. To clarify, if we have a high amount of computational power, we can decrease the tournament size. In this way,  the degree of exploration will be increased and more areas of search space will be explored. As a result, better architectures can be discovered. In contrast, for the low amount of computational power, it is good to abate the tournament size. Furthermore, due to our experiments, we hypothesis that increasing the number of generations will result in discovering better architectures. Hence, with more computational power and time, better neural architectures can be obtained. At last, it should be noted that, for multi-class image classification problems, the softmax function should be used instead of the sigmoid function in the last layer.

For future works in this area, multi-objective neural architecture search algorithms can be introduced to maximize both accuracy and precision on the validation set, and, for applications in the real-world problems, which should be done in real-time, inference time should be one of the objectives. Furthermore, new modules can be added to the search space, such as skip-connections, or more efficient search spaces can be designed to tackle problems in the real world. Additionally, further work can be done to automate the other components of the image classification process, such as data cleaning and model selection, on imbalanced datasets.

\section{Conclusions}
\label{conclusions}

We proposed a powerful and efficient algorithm termed Genetic Neural Architecture Search (GeNAS) for sperm abnormality detection. In this work, a novel fitness (objective) function named GeNAS weighting factor (GeNAS-WF) introduced to evaluate the appropriateness of each generated architecture by GeNAS.  This objective function tends to work well with all neural architecture search algorithms designed to work with datasets that are imbalanced and suffer from data shortage. Furthermore, the GeNAS algorithm can be employed to discover the best convolutional neural network architecture capable of ultimately tackling any image classification problem, specifically on noisy, low quality, and imbalanced datasets, which primarily appear in real-world scenarios. Empirically, we proved that GeNAS can ascertain better state-of-the-art architectures- in terms of accuracy, precision, and $f_{0.5}$ measure- compared to previously proposed methods, such as hand-designed CNN architectures, image processing approaches, and random search, with less amount of computational power and human effort on all three acrosome, head, and vacuole labels of MHSMA dataset. Additionally, The architectures discovered by GeNAS have exceptionally fewer parameters on the head and vacuole labels. Finally, concerning the lack of NAS research to address the challenges of real-world datasets, we recommend that further research should be done in this area of research.

\section*{References}

\bibliographystyle{model2-names}
\bibliography{Refs}

\end{document}